%% file: main.tex
\documentclass[runningheads]{llncs}
\usepackage[final,year=2026]{eccv}
\usepackage[a4paper, margin=2.5cm, columnsep=0.6cm]{geometry}

\usepackage{eccvabbrv}

\usepackage{graphicx}
\usepackage{booktabs}
\usepackage{csquotes}
\usepackage{xparse}
\usepackage{xcolor}
\usepackage[T1]{fontenc}
\usepackage{lmodern}
\usepackage{longtable}

\DeclareRobustCommand{\vidtag}[1]{{\detokenize{#1}}%
}
\DeclareRobustCommand{\function}[1]{\texttt{\detokenize{#1}}%
}
\DeclareRobustCommand{\query}[1]{%
  \texttt{\enquote{#1}}%
}

\usepackage[accsupp]{axessibility}  

\usepackage{hyperref}[pagebackref]

\usepackage{orcidlink}

\begin{document}

\title{Prompting-MammAlps: Fine-Grained Text-to-Video Retrieval for Camera-Trap Data} 

\titlerunning{Prompting-MammAlps}

\author{Valentin Gabeff\inst{1}
\and Baptiste Maquignaz\inst{1}
\and Jennifer Shan\inst{1}
\and Sepideh Mamooler\inst{1}
\and Gencer Sumbul\inst{1}
\and Blair Costelloe\inst{2,3}
\and Devis Tuia\inst{1}
\and Alexander Mathis\inst{1,4}
}

\authorrunning{V.~Gabeff et al.}

\institute{Ecole Polytechnique Fédérale de Lausanne (EPFL), Lausanne, Switzerland 
\and Max Planck Institute of Animal Behavior, Konstanz, Germany
\and University of Konstanz, Konstanz, Germany
\and \email{alexander.mathis@epfl.ch}
}
\maketitle

\begin{abstract}
Automatically retrieving videos from large camera-trap datasets remains challenging. 
Text-to-Video retrieval (TVR) methods based on large video-language models (VLMs) have potential to retrieve events of interest by describing them with simple text queries.
However, current methods often lack spatiotemporal understanding and do not generalize well to ecological data.
In this work, we introduce Prompting-MammAlps, the first camera-trap TVR benchmark, and propose a fine-grained and interpretable TVR method. 
Specifically, we trained a vision transformer to perform spatiotemporal action localization, and convert its output to structured text, describing each video. Independently, ethology-inspired queries are processed by a Large-Language Model (LLM) based coding agent to parse the structured text per video and retrieve videos accordingly. We harnessed the LLM to use functions from a custom parsing library to minimize the risk of LLM hallucinations and to improve method interpretability.
This retrieval approach applied on the Prompting-MammAlps benchmark achieved a set-based F1-score of 34\% on a test set of 135 ecologically-relevant queries and 775 candidate videos. In comparison the best zero-shot VLM achieved a F1-score of 18\%, while also lacking interpretability.

\textbf{Project page}: \href{https://cnai.epfl.ch/prompting-mammalps/}{cnai.epfl.ch/prompting-mammalps}

\end{abstract}

\begin{figure}[ht]
    \centering
-    \includegraphics[width=\linewidth]{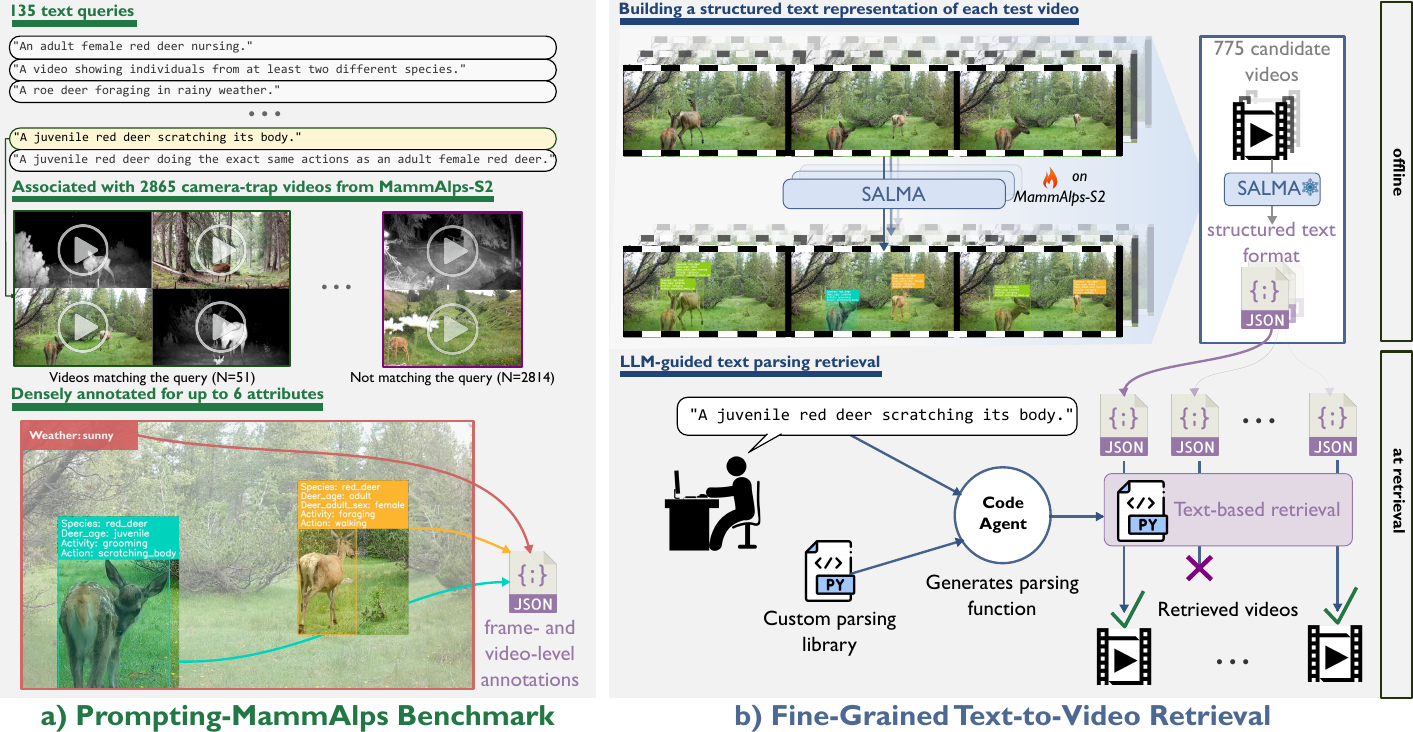}
    \caption{\textbf{Overview of Prompting-MammAlps and our TVR method.} (a) We associate a densely annotated pool of camera-trap videos with 135 queries spanning six fine-grained visual attributes and diverse spatiotemporal relationships. (b) We perform spatiotemporal action localization on the candidate videos and store each result in a structured \texttt{.json} format. For each query, we task a LLM to write a parsing function that assesses if the text representation of a candidate video matches the query or not.}
    \label{fig:overview}
\end{figure}

\section{Introduction}

Wildlife monitoring is crucial to design accurate ecosystem conservation strategies, to measure their impact over time\cite{steenweg2017scaling, kays2015terrestrial, couzin2023emerging, tuia2022perspectives}, and more generally to improve our fundamental understanding of how animals behave in complex, real-world environments \cite{anderson2014toward, couzin2009collective, koger2023quantifying,bonnetto2026behavior}. 
To achieve this, ecologists deploy networks of camera-traps \textendash among other remote sensors and other techniques\textendash ~that capture fine-grained visual information of ecological events with minimal disturbance\cite{o2011camera, rowcliffe2014quantifying, burton2015wildlife, caravaggi2017review, rovero2021camera}. However, processing this data remains time-consuming, far exceeding the time spent in data analysis\cite{young2018software, beery2019efficient, tuia2022perspectives}.
While deep learning advances have been improving camera-trap data processing, for example to filter out false positive images~\cite{beery2018recognition, hernandez2024pytorchwildlife} or to recognize species identity for some taxonomic groups\cite{rigoudy2022deepfaune}; fine-grained visual attributes (\eg animal behavior) are still typically annotated manually by experts \cite{young2018software, rovero2021camera, hofmeester2020using} or through citizen-science initiatives with notable exceptions~\cite{norouzzadeh2018automatically,liu2024joint, rodriguez2025visual, gabeff2024wildclip}.

When recording videos, camera-traps can capture rich information on animal behaviors and on social and environmental interactions~\cite{berger2011integrating, tobias2019integrating, brookes2024panaf20k, brookes2024chimpvlm, gabeff2025mammalps, vogg2025computer}. While video captures richer behavioral data, the overhead of managing and annotating it, (still) makes images the practical choice for most applications. To lift the constraints on the research questions that can be answered using camera-trap videos, the field requires better methods to automatically process and manipulate collected data. To address this, text-to-video retrieval (TVR) methods are particularly relevant. They retrieve the videos needed to address an ecological research question of interest~\cite{vendrow2024inquire, vendrow2025inquire} based on user text queries. 

The recent development of video-language models (VLMs)~\cite{wang2022internvideo, yang2023vid2seq, zhao2024videoprism, chen2024internvl, lin2024video, bardes2024revisiting} from large corpora of video-text pairs\cite{anne2017localizing, kay2017kinetics, carreira2017quo, goyal2017something, miech2019howto100m, zellers2022merlot, yang2023vidchapters} drove major advances in TVR~\cite{zhu2023deep, younis2025ai}. However, these models still fail to generalize to domain-specific data such as camera-trap imagery~\cite{dussert2024zero, gabeff2024wildclip}. 
This performance gap is in part due to: (1) the scarcity of publicly available datasets describing ecological events beyond species taxonomy; and (2) to the lack of benchmarks to properly evaluate and guide the development of large VLMs for fine-grained ecological tasks. (3)~Additionally, current TVR methods often rely on retrieving videos based on the latent similarity of few sub-sampled frames with a text prompt~\cite{zhu2023deep, younis2025ai}, which hinders interpretability and under-performs on complex prompts involving spatiotemporal reasoning~\cite{fragomeni2025leveraging}. 
While there has been a recent interest in using VLMs to describe long videos as text followed by retrieval from LLMs, the risk of LLM hallucination when describing video content or when retrieving videos given a user prompt remains an open challenge.

In this work, we tackle the above-mentioned challenges (1)-(3). First, we extend the MammAlps dataset\cite{gabeff2025mammalps} with another season of data, improve the original annotation scheme to better comply with existing ethograms, and add additional individual-level attributes for deer species (MammAlps-S2). Second, we introduce Prompting-MammAlps, the first TVR benchmark for camera-trap videos (\cref{fig:overview}a). To this end, we defined 135 open-vocabulary text queries of ecological significance, to which we exhaustively match 2865 videos from MammAlps-S2. And third, we propose a new TVR method (\cref{fig:overview}b) to improve interpretability and reasoning over spatiotemporal relationships detected in camera-trap events. In particular, we process candidate videos offline with a fine-tuned video transformer (SALMA), and represent each video's output as a structured \texttt{.json} file with frame- and video-level information. At retrieval time, we task a LLM code agent to process user queries and to output a text parsing function following a custom library of common parsing operations. The generated function is then applied to each candidate video text representation and indicates that representation's binary correspondence with the input prompt. 
As such, the code agent never interacts with the candidate videos' textual representations but only needs to translate the input prompt into a logical sequence of predefined parsing function calls, improving parsing capabilities and limiting the risk of hallucination. The interpretability of our retrieval approach lies in the intermediary frame-level description from SALMA that can be visualized on the retrieved videos, and in the generated parsing function that can be investigated by the end-user.
We evaluate our TVR method on the proposed benchmark and compare its retrieval performance with other similarity-based retrieval methods.

\label{sec:intro}

\section{Related Works}

\textbf{Text-to-video retrieval (TVR) benchmarks} have flourished in recent years given the growing need for retrieving data from large video datasets~\cite{zhu2023deep, younis2025ai}. While first benchmarks used associated captions~\cite{xu2016msr, krishna2017dense, wang2019vatex, anne2017localizing}, subtitles~\cite{lei2020tvr, miech2019howto100m} or coarse metadata~\cite{rossetto2018v3c} as a convenient source of queries, later works progressively created more realistic textual queries for retrieval, with an increased reasoning complexity~\cite{kriz2025multivent}. Since the associated videos are primarily sourced from popular multimedia platforms such as YouTube, the topic is either too broad~\cite{miech2019howto100m}, or focused on domains not relevant for ecology~\cite{zhou2018towards}. Moreover, current benchmarks do not include queries for which no video is associated with them. Together, these reasons strongly limit their use to accurately evaluate recent VLMs for TVR on camera-trap data.
Recently, Inquire~\cite{vendrow2024inquire} has become the first text-to-image retrieval benchmark applied to ecological data: a list of 250 expert-level queries was semi-automatically matched with five million images from the iNaturalist platform~\cite{van2018inaturalist, iNaturalist}. 
While there is no such benchmark for video data, several video datasets contain fine-grained visual information, which could power VLMs suited for ecological retrieval tasks~\cite{ng2022animal, chen2023mammalnet, rogers2023meerkat, kholiavchenko2024kabr, duporge2024baboonland, brookes2024panaf20k, liu2023lote, rodriguez2025visual}. Our benchmark is the first TVR benchmark for camera-trap videos and closely represents the context in which TVR could be used in practice: they span various ecological themes of interest and some queries do not have any videos associated with them.

\noindent\textbf{TVR methods} using deep learning commonly retrieve videos of high latent similarity with text queries in a joint embedding space shared between both modalities\cite{anne2017localizing, dong2019dual, gabeur2020multi, croitoru2021teachtext, luo2021clip4clip, xu-etal-2021-videoclip, dong2022reading, xue2022clip,jiang2022tencent, wang2024text}. 
This paradigm received increased attention with the development of large pretrained vision-language models, first for images~\cite{radford2021learning} and later for videos ~\cite{chen2024internvl, lin2024video, yang2023vid2seq, cicchetti2024gramian}, which can generalize well enough on some domains for zero-shot retrieval, while having a significant computational cost for the models employing cross-modal attention~\cite{song2024moviechat, lin2024video, wang2022internvideo}. 
While methods based on latent similarity or cross-modal integration apply well to images, their adaptation to videos often fails to capture long-term spatiotemporal relationships between entities and lacks explainability\cite{pulakurthi2025x}. 

\noindent\textbf{Textual representation of videos} appears as a promising solution to these issues by learning to summarize the relevant visual information of a video into a textual form~\cite{you2024toward, malik2025ravu, pulakurthi2025x,mamooler2025finetuning}.
The powerful reasoning capabilities of modern LLMs can then be used to interpret the textual representation of videos\cite{you2024toward, malik2025ravu}, or to reason on the retrieved results \cite{pulakurthi2025x}.
While not directly applied to TVR, RAVU~\cite{malik2025ravu} is a question-answering framework based on a VLM and a LLM, which jointly operate to create and parse a video spatiotemporal graph. Then, compositional reasoning on the graph is performed through Chain-of-Thought (CoT) and frames of interest to the question are retrieved by measuring the similarity between the embedded textual representations of a graph node and the grounding query.
In X-CoT\cite{pulakurthi2025x}, authors propose an interpretable re-ranking method based on LLM CoT reasoning. The LLM processes the similarity measures between videos, their structured frame-level captions generated by a VLM, and the input query. The method requires to exhaustively compute a binary preference decision for all video pairs from the top-k list, which does not scale well with k, and associated explanations from the LLM are prone to hallucinations hindering retrieval interpretability.

Our proposed Prompting-MammAlps benchmark is closest to Vendrow et al.~\cite{vendrow2024inquire}. The smaller scale of our benchmark (135 queries matched with 18 hours of video) in comparison to its internet-based counterparts is explained by the difficulty in matching query-video sets representative of fieldwork data, but comes with the advantage of high-quality annotations and domain-specific queries. Our retrieval strategy is inspired by Malik et al.\cite{malik2025ravu} and Mamooler et al. \cite{mamooler2025finetuning} when applied to TVR. It differs by constraining the LLM to perform solely prompt decomposition and parsing code generation using a predefined set of parsing functions without direct examination of the video textual representations, therefore limiting hallucination and improving interpretability.

\section{Prompting-MammAlps}
\subsection{Extending MammAlps: MammAlps-S2}
\label{sec:data}

We extended the initial MammAlps dataset~\cite{gabeff2025mammalps} (i) by adding another season of densely annotated data (acquired in 2024), (ii) by annotating two new attributes describing deer individuals status for both years, and (iii) by improving the ecological validity of the annotation scheme. Specifically, we:

(i) replicated the original data collection protocol by deploying three cameras in the same three sites as in 2023. While the position of the cameras in Site 2 were closely matched, the position of the Site 1\&3 cameras required adjustment and did not share the same field-of-view as the ones placed in the first season. We therefore refer to these cameras as C4-C6 (instead of C1-C3). Cameras were deployed in the field for 10 weeks (Jun-Aug 2024) recording at day and night (IR illumination). This period does not overlap with the previous data acquisition period of MammAlps (Aug-Oct 2023; see Supp. Mat.~\ref{app-sec:data}), adding behavioral and morphological diversity to the dataset, as well as increasing the number of samples for rare species (\ie mountain hare). 

(ii) improved the annotation scheme for ungulate species by collecting and integrating behavior descriptions from existing ethograms \cite{prikhod2011behavior, cap2002phylogeny}. 

(iii) increased label diversity by manually annotating deer age as a binary category (\textit{juvenile} or \textit{adult}) based on the apparition of sexual dimorphism traits, along with sex for adult individuals (\textit{male} or \textit{female}). We focused on deer as they are in the majority of events. Furthermore, we manually refined the dense annotations of the videos from the 2023 dataset as an opportunity to improve the labeling quality and to use a common label space for both seasons. 

We refer to this extended dataset as \textit{MammAlps-S2}. MammAlps-S2 comprises of 2865 videos totaling 18 hours of high-resolution recording acquired from 15 different camera views. The videos were densely annotated such that every animal contains frame-level annotations for its behavior (up to two low-level actions and one high-level activity), individual-level attributes (species, age for deer species, sex for adult deer species), and weather at the video-level. The annotation process yielded 3227 individual tracks (avg. duration of 15.6s), 1.5M bounding boxes, and an average of 1.13 individuals per video.
We included a small proportion of false positive (\ie absence of visible vertebrates) videos since these are common in camera-trap datasets. 

We randomly split camera-trap acquisition days in two groups to define a train and a test video set while maintaining a similar label distribution across sets. The train and test sets contain 2090 and 775 videos, respectively. We further detail the final MammAlps-S2 dataset in Supp. Mat. \ref{app-sec:data} and illustrate the distribution of attributes across video splits in \ref{app-sec:video-split}.

\subsection{Text-to-Video Retrieval Benchmark}
\label{sec:queries}

\begin{figure}[ht!]
    \centering    \includegraphics[width=\linewidth]{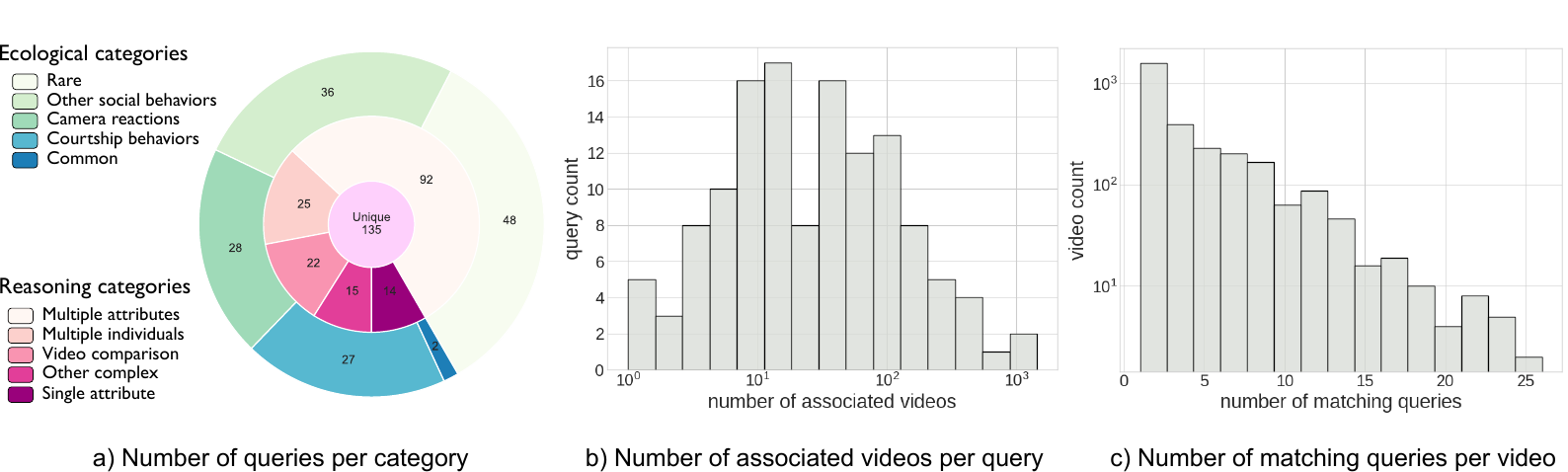}
    \caption{\textbf{Prompting-MammAlps Benchmark}. (a) The 135 queries are categorized in five ecological and five reasoning categories. One query can belong to multiple categories. (b-c) Number of matching videos per query and inversely (excluding zero entries). }
    \label{fig:benchmark}
\end{figure}

To create the TVR benchmark, we identified 135 text queries in collaboration with a behavioral ecologist (B.C.).
These queries span four ecological categories: \textsl{courtship} related behaviors, \textsl{other social behaviors}, \textsl{camera reactions}, and any other \textsl{rare} events (\cref{fig:benchmark}A). We additionally included a \textsl{common} category designed to retrieve (\ie filter out) the most common events (\eg\query{A red deer foraging only.}). 
We also categorized queries based on their reasoning complexity. \textsl{Single} and \textsl{multi-attributes} queries require processing information from one, and two or more different attributes (\eg\query{A juvenile red deer suckling.}). \textsl{Multiple individuals} queries ask information about two or more individuals (\eg\query{A video of two or more wolves.}). \textsl{Complex} queries require more reasoning than checking for the presence of some attributes (\eg\query{An animal reacting to a camera and then foraging.}). 
We also propose \textsl{video comparison} queries which require integrating information from a reference video with the other candidate videos (\eg\query{An individual doing the same sequence of actions as the individual of \vidtag{S2_C1_E323_V0074}.}). 

We used the ground-truth annotations to semi-automatically match the camera-trap videos of MammAlps-S2 to each query. Only videos containing at least a segment fully matching the elements from a query were associated with it, while incomplete matches and non-relevant videos were excluded from this list. 
We considered the training set of Prompting-MammAlps as the aforementioned list of queries associated with the train videos. Similarly, the test videos, referred to as candidate videos, were associated with the same list of queries to constitute the test retrieval set.

On average, queries are matched with 78.6 videos (min.: 0; max.: 1439). Seven queries (13 when considering the test set) do not have any video associated with them but are still kept in the dataset as a user may look for events that were never captured by any camera (\cref{fig:benchmark}B). Single videos can also be matched to multiple queries, on average to 3.7 of them (min.: 0; max.: 26). Similarly, some videos are not associated with any query (\cref{fig:benchmark}C). Additional benchmark statistics can be found in Supp. Mat.~\ref{app-sec:benchmark}.

\section{Fine-grained text-to-video retrieval strategy}
Our end-to-end TVR strategy processes the textual user query and the candidate videos separately. This enables to process candidate videos offline prior to retrieval, improving computational efficiency and speed at retrieval. 

\subsection{SALMA: Spatiotemporal Action Localization for MammAlps}

\begin{figure}[ht]
    \centering
    \includegraphics[width=\linewidth]{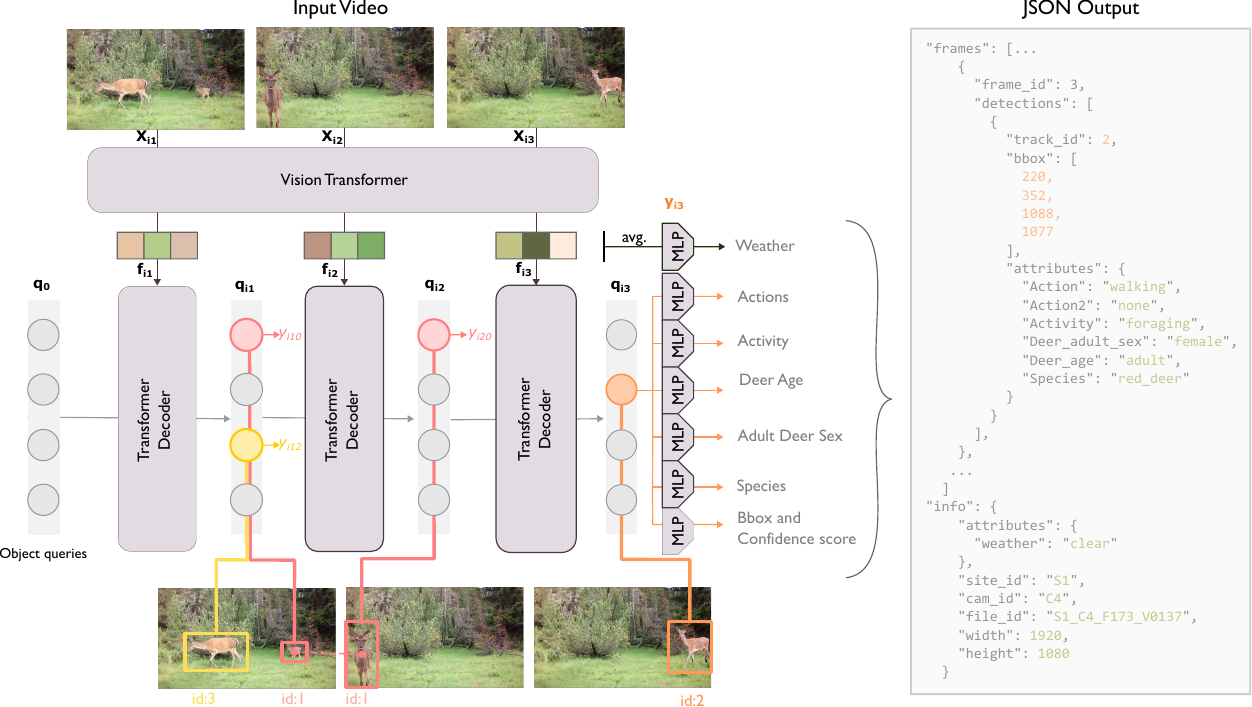}
    \caption{\textbf{From raw videos to their structured text representations.} SALMA progressively decodes object queries over a sequence of frames and outputs a track activation score for each object query $q_{i,t,j}$ (shown in color if above the activation threshold $\sigma_{s}$) and its associated list of predictions $y_{i,t,j}$.
    We created a \texttt{.json} file containing video- and frame-level information given the predicted attributes of the activated object queries.}
    \label{fig:salma}
\end{figure}

To process candidate videos, we aim to exhaustively extract the visual information that can be useful to retrieve videos matching the queries. This involves tracking individuals across frames, performing individual-level attributes recognition and action segmentation on each detected animal track. For this, we used a monolithic architecture adapted from DETR\cite{carion2020end} and MOTR\cite{zeng2022motr} (\cref{fig:salma}). We trained an encoder-decoder transformer model to progressively decode object queries over a sequence of frames $\textbf{X}_{i,t}, i \in [0, N_{v}], t\in[0, N_{f,i}]$ where $N_v$ is the number of videos and $N_{f,i}$ the number of frames for video $i$, and used them to predict both frame- and individual-level attributes ${\textbf{y}_{i,t}}$. The encoder (parametrized by $\theta_{en}$) follows the vision transformer architecture from VideoMAE~\cite{tong2022videomae, wang2022internvideo,gabeff2025mammalps} and outputs a sequence of video tokens ${\textbf{f}_{i,t}}$. The transformer-based decoder (parametrized by $\theta_{dec}$) progressively decodes learnable object queries $\textbf{q}_{i,t}$ (initialized at the beginning of every frame sequence to $\textbf{q}_0$) through cross-attention with ${\textbf{f}_{i,t}}$. Each object query $j$ is given as input to six multi-layered perceptions (MLP) heads, predicting bounding-box coordinates and an activation score $s_{i,t,j}$, species identity, actions, activity, deer age, and adult deer sex categories, respectively. Differently from Carion et al.~\cite{carion2020end}, we did not add a null-object class to indicate if an object query $q_{i,t,j}$ is not associated with any detection. Instead, we applied a fixed threshold $\sigma_s$ on $s_{i,t,j}$ to indicate whether $q_{i,t,j}$ is associated with an animal track ($s_{i,t,j}\geq\sigma_s$) or if it is an inactive query ($s_{i,t,j}<\sigma_s$).

During training, we learned the parameters of SALMA in two curriculum steps. Firstly, we learned $\theta_{en}$, $\theta_{dec}$, $q_0$, and the MLP head parameters used to predict bounding box coordinates and the query activation scores $s_{i,t,j}$. The matching between ground-truth animal detections and each object queries is done through Hungarian matching based on association costs~\cite{carion2020end}. We applied an $L_1$ loss and a $gIoU$ loss on the bounding box coordinates, and used binary-cross entropy to learn $s_{i,t,j}$. To maintain the temporal consistency of object queries, we computed an InfoNCE contrastive loss where pairs of object queries $q_{i,t,j}$ and $q_{i, t+1, j}$ are considered as positives, while any other pair at this time step in the video and in the batch are considered as negatives~\cite{manasyan2025temporally}. Secondly, we learned the remaining MLP heads parameters using categorical cross-entropy, while still optimizing for the parameters of the first curriculum step. 

At inference, we ran SALMA on every frame of a candidate video, propagating object queries over frame sequences. To account for the input resolution ratio difference between training (1:1) and inference (16:9), we ran inference on two square-cropped views of the image, and used Hungarian matching to match predictions across both views. For individual- and video-level attributes, we took the most commonly predicted value over predicted tracks and the entire video, respectively. The output of each decoding step was collected in a \texttt{.json} text file: for every frame $t$, we wrote a list of detection objects, each with an associated track id $j$ corresponding to the activated object query $q_{i,t,j}$, and containing a list of attributes corresponding to the predicted labels $y_{i,t,j}$. The weather attribute was added in the \enquote{info} section of the file. 

\subsection{LLM-guided text parsing}

To process text queries, we translated each query into an executable function that operates on a video’s \texttt{.json} representation and returns a boolean value indicating whether the annotation satisfies the query’s semantic constraints.
We leveraged a LLM to synthesize these functions from natural language queries.
Specifically, we designed a library of 23 functions that operate over structured video annotations. These primitive functions can be combined to form executable predicate functions without having to input the content of the \texttt{.json} file to the LLM. They can be used to check for the presence of some specific attributes (\eg\function{check_tracks_contains_species(tracks, species_name)}, or to extract general information (\eg\function{get_unique_activities(tracks)}) about a given track. We list them in Supp. Mat.~\ref{app-sec:library}.
The library specification was provided to the LLM as a prompt together with the query, the possible labels, and three in-context reasoning examples from unseen queries, enabling the LLM to generate a predicate function composed of the available primitives.
The resulting predicate functions were then executed on every candidate video's text representation. If the conditions stated in the function were all met for a given \texttt{.json} file, we considered it a match to the query and returned the associated video to the user.

\section{Experiments}

To evaluate our approach on Prompting-MammAlps, we first trained SALMA on the 2090 training videos for 500 epochs in each curriculum learning step. We used a large pretrained vision transformer encoder backbone~\cite{wang2022internvideo}. At inference, we processed the 775 candidate test videos using a temporal stride of four frames and applied forward-filling to derive missing entries. Additional architecture and training details can be found in Supp. Mat. \ref{app-sec:salma}.

We used smolagent~\cite{smolagents} for the LLM-based coding framework and evaluated the prompt interpretation and parsing code generation capabilities of four different LLMs: Qwen3-8B~\cite{yang2025qwen3}; Llama-3.1-8B-Instruct~\cite{llama3modelcard}; Mistral-7B-Instruct~\cite{jiang2023mistral}, and Apertus-8B-Instruct~\cite{swissai2025apertus}. This framework allows for efficient multi-step reasoning, since any deviation from the list of provided parsing functions or from the possible attribute labels results in an error that is forwarded to the code agent, initiating a new trial. The agent finished its execution once it successfully calls the generated function on an example \texttt{.json} file without any error, or if it failed to generate an executable function after more than ten trials.

We compared our approach against zero-shot evaluation of CLIP4Clip~\cite{luo2021clip4clip}, GRAM~\cite{cicchetti2024gramian} and, InternVideo2.0 1B~\cite{wang2022internvideo}. Since Vendrow et al. \cite{vendrow2024inquire} found SigLIP SO400m-14\cite{zhai2023sigmoid} to be the most effective retrieval model, we also applied this backbone to CLIP4Clip, to which we refer as SigLIP4Clip. To better compare with our supervised approach, we fine-tuned CLIP4Clip on query-video pairs from the train set of query-video associations. We provide further details for these experiments in Supp. Mat. \ref{app-sec:sota}. 

Since the methods used as comparison provide a ranking based on a distance~\cite{cicchetti2024gramian} or a similarity score~\cite{luo2021clip4clip, wang2022internvideo} between the textual and the visual modality, they are not directly comparable to ours. Our approach does not compute a relevance score but only a binary matching decision for every candidate video and query pair and we cannot compute standard ranking-based metrics such as mAP or Recall@k. Instead, we computed the F1-score between the retrieved set of candidate videos and the ground-truth set for every query. For queries not associated with any ground-truth video, we computed the F1-score based on the filtered set (\ie what is left after retrieval). 
We then averaged each F1-score over the list of queries. 
For the comparison baselines, we computed the optimal similarity decision threshold for each individual query on the training set, which contains the same queries but different videos, and then computed the F1-score of the binarized output using this optimized decision threshold. Additionally, we cannot compute the performance of these methods on queries requiring a comparison with a reference video as it requires a two-steps reasoning process, and we therefore exclude these queries from the comparative analysis.

\section{Results}
\subsection{End-to-end retrieval}

We report the end-to-end retrieval performance of our approach and of the zero-shot performance of comparative VLMs (\cref{tab:retrieval}). Since our method is bounded by the capabilities of the code agent to generate correct parsing functions, we report the best possible retrieval performance obtained when the generated parsing functions are directly applied to the ground-truth annotations (row Agent $+$ Oracle). When applying the generated functions to the output from SALMA instead of the ground-truth (row Agent$+$SALMA), we obtained an averaged F1-score of 0.34, suggesting that SALMA does not reliably extract the visual information necessary for accurate retrieval.  For the reasoning categories, our approach was most effective on simple queries referring to a single visual attribute, and performed comparatively lower on queries involving multiple individuals.

VLMs zero-shot retrieval performances were all lower than our proposed method, highlighting poor generalization to this ecological domain. To bridge this domain gap, we fine-tuned CLIP4Clip on MammAlps-S2 by using the training query-video pairs, which led to a consistent gain in performance across categories. However, this naïve supervised approach still underperformed in comparison to ours, suggesting that more complex retrieval strategies are required to perform well on the proposed benchmark (\cref{tab:retrieval}). The comparison remains limited since the coding agent has access to common reasoning operations through the parsing library.

When comparing the four LLM performances on this task (\cref{tab:llms}), we observed that Qwen3-8B yielded the best generative capabilities with a F1-score of 0.87, and we therefore used this model for the remainder of the study.

\begin{table}[ht]
    \caption{\textbf{F1-Score retrieval performance on Prompting-MammAlps}. $\dagger$: fine-tuned on MammAlps-S2, zero-shot evaluation otherwise; n/a: VLMs performance cannot be evaluated on queries referencing an additional video (Vid. Comp.). 
    }
    \resizebox{\columnwidth}{!}{%
    \begin{tabular}{l | ccccccc}
    \toprule
    \textbf{\begin{tabular}[c]{@{ }l@{ }}Ecological Cat. \\ (Support)\end{tabular}} &
    \textbf{\begin{tabular}[c]{@{ }c@{ }}All\\ (135)\end{tabular}} &
    \textbf{\begin{tabular}[c]{@{ }c@{ }}w/o Vid. Comp.\\ (113)\end{tabular}} &
    \textbf{\begin{tabular}[c]{@{ }c@{ }}Rare\\ (48)\end{tabular}} &
    \textbf{\begin{tabular}[c]{@{ }c@{ }}Courtship\\ (27)\end{tabular}} &
    \textbf{\begin{tabular}[c]{@{ }c@{ }}Other Social\\ (36)\end{tabular}} &
    \textbf{\begin{tabular}[c]{@{ }c@{ }}Cam. Reaction\\ (28)\end{tabular}} &
    \textbf{\begin{tabular}[c]{@{ }c@{ }}Common\\ (2)\end{tabular}} \\
    \midrule
    Agent + Oracle & 0.87 & 0.89 & 0.94 & 0.79 & 0.90 & 0.82 & 0.98 \\
    \midrule
    CLIP4Clip \cite{luo2021clip4clip} &   n/a  & 0.17  & 0.16  & 0.13 & 0.26  & 0.14 & 0.40   \\
    SigLIP4Clip                       &   n/a  & 0.17 &  0.17 & 0.12  & 0.25  & 0.12  & 0.37  \\
    GRAM  \cite{cicchetti2024gramian} &   n/a  & 0.16  & 0.16  & 0.08  & 0.25  & 0.12  &  0.35 \\
    InternVideo2.0 \cite{wang2022internvideo}  &   n/a  &  0.18  & 0.16 & 0.15 &  0.28   &  0.12   &  0.40  \\
    CLIP4Clip$\dagger$ &  n/a  & 0.26 & 0.29 & 0.22  &  0.30 &  0.18 & 0.70 \\
    Agent + SALMA$\dagger$  & \textbf{0.34} & \textbf{0.37} & \textbf{0.43} & \textbf{0.34} & \textbf{0.35} & \textbf{0.21} & \textbf{0.87} \\
    \midrule
    \textbf{\begin{tabular}[c]{@{ }l@{ }}Reasoning Cat.\\ (Support)\end{tabular}} &
    \textbf{\begin{tabular}[c]{@{ }c@{ }}All\\ (135)\end{tabular}} &
    \textbf{\begin{tabular}[c]{@{ }c@{ }}w/o Vid. Comp.\\ (113)\end{tabular}} &
    \textbf{\begin{tabular}[c]{@{ }c@{ }}1 attr.\\ (14)\end{tabular}} &
    \textbf{\begin{tabular}[c]{@{ }c@{ }}2+ attr.\\ (92)\end{tabular}} &
    \textbf{\begin{tabular}[c]{@{ }c@{ }}2+ indiv.\\ (25)\end{tabular}} &
    \textbf{\begin{tabular}[c]{@{ }c@{ }}Vid. Comp.\\ (22)\end{tabular}} &
    \textbf{\begin{tabular}[c]{@{ }c@{ }}Complex\\ (15)\end{tabular}} \\
    \midrule
    Agent + Oracle & 0.87 & 0.89 & 1.00  & 0.89 & 0.80 & 0.75 & 0.69 \\
    \midrule
    CLIP4Clip \cite{luo2021clip4clip} &   n/a  & 0.17  & 0.12  & 0.18 & 0.13  & n/a  & 0.13   \\
    SigLIP4Clip                     &   n/a  & 0.17 & 0.13  & 0.18  & 0.10  & n/a  & 0.14  \\
    GRAM  \cite{cicchetti2024gramian} &   n/a  & 0.16  & 0.11  & 0.17  & 0.11  & n/a  & 0.13  \\
    InternVideo2.0 \cite{wang2022internvideo}  &  n/a   &  0.18   & 0.10 & 0.19 & 0.15 &  n/a   & 0.15   \\
    CLIP4Clip$\dagger$  & n/a  & 0.26 & 0.29 & 0.25 & 0.20 &  n/a   & 0.25  \\
    Agent + SALMA$\dagger$  & \textbf{0.34} & \textbf{0.37} & \textbf{0.44} & \textbf{0.37} & \textbf{0.23} & \textbf{0.19} & \textbf{0.26} \\
    \bottomrule
    \end{tabular}%
    }
    \label{tab:retrieval}
\end{table}

\begin{table}[ht]
    \caption{\textbf{Oracle performance comparison between LLM code agents}. Qwen3-8B is the model yielding the best retrieval performance when applying generated functions to the ground-truth \texttt{.json} representation of candidate videos.}
    \centering
    \begin{tabular}{l c}
    \toprule
       \textbf{LLM} & \textbf{F1-score} \\
       \midrule
        Mistral-7B-Instruct \cite{jiang2023mistral} + Oracle & 0.38 \\ 
        Apertus-8B-Instruct \cite{swissai2025apertus} + Oracle  & 0.41 \\ 
        LLama-3.1-8B-Instruct \cite{llama3modelcard} + Oracle  &   0.42 \\ 
        Qwen3-8B \cite{yang2025qwen3}+ Oracle  & \textbf{0.87} \\ 
        \bottomrule
    \end{tabular}
    \label{tab:llms}
\end{table}

\subsection{SALMA}
\label{sec:results-salma}
We analyzed the performance of SALMA by first computing its multi-object tracking capabilities after the first (SALMA-MOT) and the second (SALMA) curriculum steps (\cref{tab:mot-f1}a). As a comparison, we report the tracking performance obtained on the candidate videos when using MegaDetector v5a~\cite{hernandez2024pytorchwildlife} for detection, followed by ByteTrack \cite{zhang2022bytetrack} for tracking. While this latter approach gave the highest tracking accuracy (HOTA) and the least number of identity swaps, the detection performance of SALMA-MOT was superior and led to the best IDF1 score, while being end-to-end and with minimal post-processing steps. The higher detection accuracy might be explained by the fact that our approach is directly fine-tuned on MammAlps-S2, while MegaDetector\cite{hernandez2024pytorchwildlife} has been trained on a vast set of camera-trap datasets, but excluding the alps. 
The tracking performance of SALMA-MOT slightly degraded after the second curriculum step (SALMA) as we add more learning objectives. Removing the contrastive objective on the object queries $\textbf{q}_{i,t}$ also decreased the tracking performance of SALMA-MOT while yielding a relatively similar detection accuracy. We therefore argue that this loss is useful to maintain temporal consistency in the context of multi-object tracking, as it was previously demonstrated for object-centric learning\cite{manasyan2025temporally}. Yet SALMA does not have an explicit object tracking learning objective based on the track ID as a memory-based model would have\cite{zeng2022motr, carion2025sam}. Both objectives are compatible and could be explored to improve the performance of SALMA.

Second, we report the behavior segmentation and multi-attributes recognition performance of SALMA after the second curriculum step (\cref{tab:mot-f1}b). The averaged F1-scores per attribute were within the same range of previously reported results from the first benchmark of MammAlps~\cite{gabeff2025mammalps}, while the task was considerably simpler (\ie multi-attribute classification from pre-processed clips). 

\begin{table}[t]
\caption{\textbf{Tracking and attribute performance}. (a) SALMA-MOT was trained for multi-object tracking only. Removing the contrastive loss between the object queries $\textbf{q}$ decreases tracking quality. (b) Attributes were predicted after the MOT step. False positive tracks that were not matched to any ground-truth predictions also counted as false positive attribute predictions, and conversely for false negative tracks.}
\centering

\begin{subtable}{\linewidth}
\centering
\setlength{\tabcolsep}{8pt}
\begin{tabular}{l c c c c}
\toprule
\textbf{Model} & HOTA$\uparrow$ & IDF1$\uparrow$ & DetA$\uparrow$ & ID$_{swp}\downarrow$ \\
\midrule
SALMA-MOT w/o InfoNCE loss & 67.5 & 75.4 & 64.9 & 3485 \\
SALMA-MOT & 72.9 & \textbf{84.5} & \textbf{67.7} & 192 \\
SALMA & 69.0 & 81.1 & 62.6 & 202 \\
MegaDetector\cite{hernandez2024pytorchwildlife} $\rightarrow$ ByteTrack\cite{zhang2022bytetrack} & \textbf{76.2} & 81.4 & 67.2 & \textbf{93} \\
\bottomrule
\end{tabular}
\caption{Multi-object tracking performance}
\end{subtable}

\begin{subtable}{\linewidth}
\setlength{\tabcolsep}{6pt}
\centering
\begin{tabular}{l c c c c c c}
\toprule
\textbf{Model}  & Species  & Activity  & Actions  & Deer Age  & Adult Deer Sex  & Weather  \\
\midrule
SALMA & 0.48 & 0.40 & 0.29 & 0.70 & 0.75 & 0.76 \\
\bottomrule
\end{tabular}
\caption{Per-frame F1-score (macro-avg.) per attribute category}
\end{subtable}
\label{tab:mot-f1}
\end{table}

\subsection{Agent-based prompt interpretation}

We report ablation results on the context provided to the Qwen3-8B agent (\cref{tab:ablation-llm}).
The vanilla agent did not have access to any parsing function. We still provided it a template of the \texttt{.json} structure and a description of the possible labels, as the task would have been unnecessarily complicated otherwise. Adding the custom parsing library had the strongest effect on retrieval performance ($+0.34$ F1-score), which is expected as the logic to extract complex information such as action sequences is error-prone for the LLM. 
We added explicit constraints on the label space in the last row, by raising an error and returning it to the agent when it calls elements not belonging to the available label list. While it provided minimal performance gain ($+0.03$), it proved to be useful to disambiguate between \enquote{Actions} and \enquote{Activities}.

\begin{table}[]
    \setlength\tabcolsep{14pt}
    \caption{\textbf{Ablation of the LLM framework components using ground-truth annotations}. The last row corresponds to the Agent $+$ Oracle row in \cref{tab:retrieval}.}
    \centering
    \begin{tabular}{l l }
       \toprule
       \textbf{Qwen3-8B}  & \textbf{F1-score (gain)} \\
        \midrule
        vanilla & 0.44 ($ - $)  \\ 
        + parsing library & 0.78 ($+0.34$) \\ 
        + three in-context examples & 0.84  ($+0.06$)  \\ 
        + explicit constrain on possible labels & 0.87  ($+0.03$)  \\ 
        \bottomrule
    \end{tabular}    
    \label{tab:ablation-llm}
\end{table}

\subsection{Retrieval process decomposition}

\begin{figure}[ht!]
    \centering
    \includegraphics[width=\linewidth]{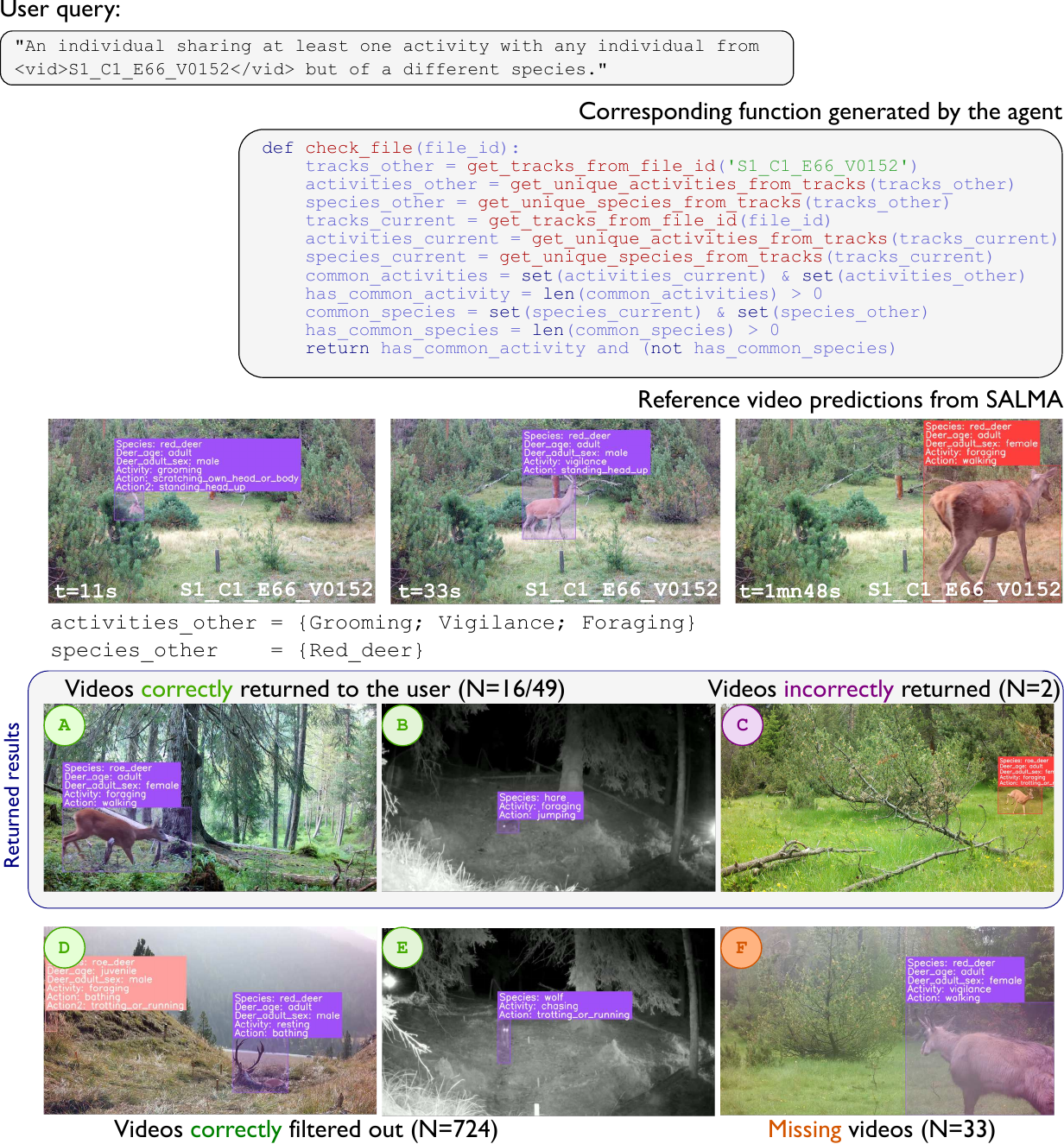}
    \caption{\textbf{Retrieval process decomposition.} For illustrative purposes, we chose the "no label constrain" version of Qwen3-8B (third row from \cref{tab:ablation-llm}) which produced a concise code but with a logic error that we discuss.
    Example frames are shown; numbers in parentheses indicate video counts. }
    \label{fig:decomposition}
\end{figure}

To illustrate our approach, we show a step-by-step retrieval process for the following user query: \query{An individual sharing at least one activity with any individual from \vidtag{S1_C1_E66_V0152} but of a different species.} (\cref{fig:decomposition} and Supp. Mat. \ref{app-sec:retrieval}). The generated code from the selected LLM agent correctly extracted the activities and the species from both the reference video and each candidate video, but failed to compare them at the individual level, which is a first source of retrieval error. We then visualized predictions from SALMA on the reference video. The extracted list of activities contained \textsl{grooming}, which is a false prediction. 
Videos A and B are correctly returned, despite the inexact LLM's logic and some attribute misclassifications. Video C is a false positive because the \textsl{roe deer} is not \textsl{foraging}. 
Videos D and E are correctly filtered out, although if the parsing function logic was correct, video D would have been returned to the user as it contains a false positive \textsl{roe deer} detection \textsl{foraging}. The model failed to return Video F as it detected a \textsl{red deer} instead of a \textsl{chamois}.

This illustrative process shows the interpretability of our TVR method. Not only can the user visualize intermediary predictions from SALMA, but they can also understand how the LLM interpreted their query through examination of the generated function, and potentially refine their request accordingly.

\section{Discussion}

The retrieval performance of baselines demonstrates that our benchmark poses difficulties to current open-source VLMs that process a relatively small number of frames per video. 
Using an agent-based strategy, we show that with an oracle tracking, segmentation and classification model, one can achieve high retrieval performance. The lower performance obtained when using SALMA's predictions therefore suggests that extracting complex visual information and linking it to natural language remains the performance bottleneck more than domain-specific language understanding in this context. 
The benchmark design also meets some unaddressed needs such as the ability to evaluate performance on queries not matched to any video, or to integrate information from an additional reference video.
Overall, Prompting-MammAlps is a unique TVR benchmark which translates needs from ecologists into a benchmark to better evaluate advances in domain-specific video-language understanding. 

Unlike most common TVR benchmarks, Prompting-MammAlps is not ranking-based but operates on a set of associated videos per query. Our methodology could be extended to a ranking-based retrieval strategy by returning a similarity score between a user query and a given \texttt{.json} file instead of the binary association. This would allow to answer more complex queries containing continuous information such as \query{A juvenile following a hind.}, and would constitute a natural extension. The benchmark design could also be extended by leveraging the available audio modality, as it is done in other related TVR works~\cite{cicchetti2024gramian, kriz2025multivent}.

Beyond our proposed benchmark, the dense tracking annotations from the MammAlps-S2 dataset can be further exploited to create other multi-modal benchmarks. For example, one could use them to automatically generate video captions or visual question-answer pairs without requiring to explicitly process the videos (see Supp. Mat. \ref{app-sec:extensions}). The dense tracking annotations can also benefit method development and evaluation for animal detection and tracking in complex environments with frequent occlusions.

\section{Conclusion}

We propose Prompting-MammAlps, a TVR benchmark containing 18h of high-resolution wildlife camera-trap recordings densely annotated at the frame-level matched to 135 expert-level queries. This specific visual and semantic domain brings unique challenges that are not well addressed by current VLMs. Our analysis suggests that perception rather than reasoning in VLMs is the biggest bottleneck to solve the benchmark. We hope that the present study will serve as inspiration to keep fostering development in video-language modeling and understanding, with the ultimate objective of supporting biologists in data management, processing and analysis.

\section*{Acknowledgments}

We thank members of the Mathis Group for Computational Neuroscience \& AI (EPFL) and of the Environmental Computational Science and Earth Observation Laboratory (EPFL) for their feedback and fieldwork efforts. 
We also thank members of the Swiss National Park monitoring team for their valuable support and feedback. The project was approved by the Research Commission of the National Park.
This project was partially funded by EPFL’s SV-ENAC I-PhD program (G.V.), Swiss SNF grant (320030-227871), and the Horizon Europe grant 101213369 (DVPS).

\section*{Author contributions}

V.G., D.T., and A.M. wrote the manuscript with input from all authors; V.G., D.T. and A.M. collected the data; V.G. and J.S. processed and annotated the data; B.C. and V.G. designed the queries. B.M., V.G. and G.S. developed SALMA; S.M.and V.G. worked on LLM-guided parsing retrieval framework. V.G. and J.S. ran baselines. D.T. and A.M. supervised the project and acquired funding.

\input{main.bbl}

\input{supplementary_material}

\end{document}

%% file: supplementary_material.tex
\newpage

\renewcommand{\thesection}{S\arabic{section}}
\renewcommand\thefigure{S\arabic{figure}}
\renewcommand\thetable{S\arabic{table}}

\setcounter{section}{0}
\setcounter{figure}{0}
\setcounter{table}{0}

\section{Supplementary Material}

Here we provide further details on MammAlps-S2 and the annotation scheme in \cref{app-sec:data}. \cref{app-sec:video-split} contains more statistics about the train and test video sets. We list the queries and their associated videos in \cref{app-sec:benchmark}. Details about SALMA's architecture, training and inference strategies, and F1-score results per attribute category can be found in \cref{app-sec:salma}. Our experimental setup to evaluate text-to-video retrieval (TVR) Vision-Language Models (VLMs) on Prompting-MammAlps is detailed in \cref{app-sec:sota}. The end-to-end retrieval per-query performances of Agent+Oracle, Agent+SALMA, CLIP4Clip\cite{luo2021clip4clip}, and InternVideo2.0\cite{wang2022internvideo} are listed in \cref{app-sec:per_query}. The library of parsing functions available to the code agent is developed in \cref{app-sec:library}. We illustrate two additional end-to-end retrieval results in \cref{app-sec:retrieval}. In \cref{app-sec:extensions}, we discuss how MammAlps-S2 can be used as a source dataset to build other vision-language tasks.

\subsection{Details on MammAlps-S2}
\label{app-sec:data}

\subsubsection{Updated annotation scheme.}
MammAlps-S2 combines two seasons of camera-trap video data acquired in the Swiss National Park, in summers 2023 and 2024. We refer the reader to~\cite{gabeff2025mammalps} for details on data collection and data processing; we note that we had accidentally misreported the study period for 2023, it went from late August to early October (\cref{app-fig:attributes_per_split}). MammAlps-S2 contains observations for seven species: red deer (\textit{Cervus elaphus}), roe deer (\textit{Capreolus capreolus}), red fox (\textit{Vulpes vulpes}), wolf (\textit{Canis lupus}), mountain hare (\textit{Lepus timidus}), chamois (\textit{Rupicapra rupicapra}), and marten (\textit{Martes martes}). Chamois and martens were absent from the original MammAlps dataset.

As part of MammAlps-S2, we updated the annotation scheme for ungulate species based on two previously defined ethograms \cite{cap2002phylogeny, prikhod2011behavior}. In \cite{cap2002phylogeny}, the authors explored the relationship between phylogeny and behavior in several species of \textit{Cervidae}, which includes the roe deer and red deer species, and compared it against other taxonomic families such as \textit{Bovidae}, which includes the chamois species. In \cite{prikhod2011behavior}, the authors clustered behavioral patterns and relate them to phylogenetic relationships among families from the \textit{Artiodactyla} order, which includes the three aforementioned species. The first study described 87 behavioral \enquote{characters}, and the second 45 behavioral \enquote{features}. Both lists of behaviors were integrated to define two lists of 27 actions and 13 activities (following the nomenclature from \cite{anderson2014toward,stoffl2025elucidating} and \cite{gabeff2025mammalps}): 
18 behaviors found a direct equivalence in the previous scheme and were minimally adjusted; 63 behaviors were excluded because they were never observed (\eg behaviors pertaining to species absent from the dataset); 23 behaviors were additionally excluded as their interpretation was difficult to assess given the limited context provided by camera-traps (\eg \textit{presence of patterns of territorial behavior}), or because they represented too fine-grained motions (\eg \textit{leaving the ground by raising the front part of the body}) that are beyond the scope of this study; 27 behaviors were grouped into higher-level categories (\eg the different ways of shaking head or any other body parts are grouped into \textit{shaking head or body}). 
After label refinement of the original MammAlps dataset, the activity label was updated for 10\% of the frames containing individuals, the first action for 6\%, the second action for 3\% and the species for 1\%.
Four actions and one activity were not retained in the final dataset because they were supported by a single observation only. The final lists of selected activities and actions are the following:
\begin{itemize}
    \item \textbf{Activities}: \textsl{Foraging; Vigilance; Unknown; Camera reaction; Grooming; \newline Courtship; Marking or wallowing; Resting; Escaping; Chasing; Playing}; and \textsl{Nursing.}
    \item \textbf{Actions}: \textsl{Walking; Standing head up; Standing head down; Trotting or running; Grazing; Unknown; Sniffing; Looking at a camera; Scratching own head or body; Vocalizing; Jumping; Rubbing antlers on the ground; Shaking head or body; Drinking; Laying down; Pawing the ground; Bathing; Browsing; Defecating; Stretching body; Urinating; Preparing to suckle;} and \textsl{Suckling.}
\end{itemize}
\textsl{Unknown} annotations correspond to behavior segments that cannot reliably be annotated given the available context (including audio), for example in case of partial occlusion. 

\subsubsection{Cross-season comparison.}

\begin{figure}
    \centering
    \includegraphics[width=\linewidth]{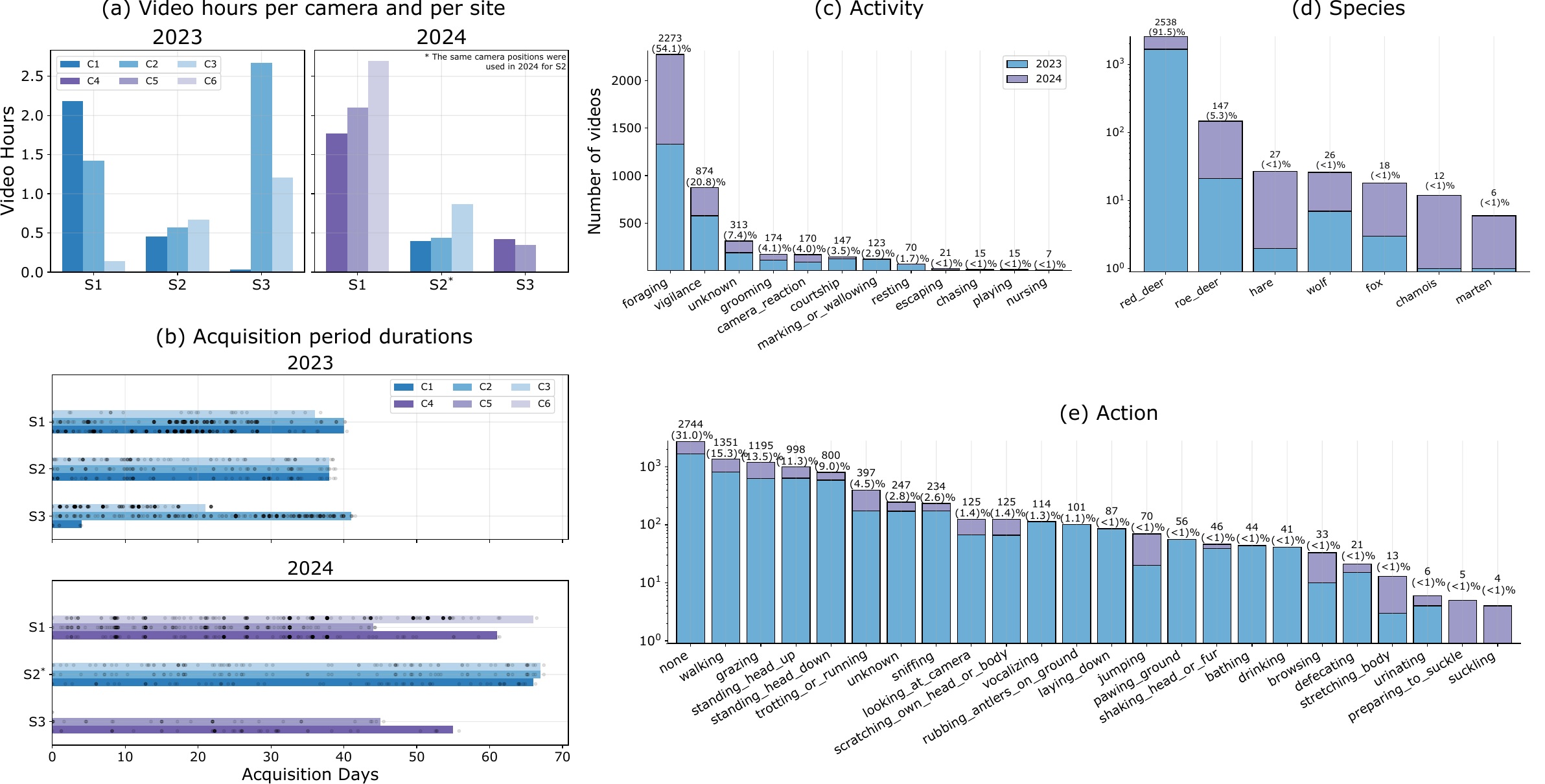}
    \caption{\textbf{Comparison between 2023 and 2024 seasons}. Taken together, both seasons constitute the MammAlps-S2 dataset. (a) Number of video hours per camera and per acquisition site. Three cameras were placed at each site. (b) Number of acquisition days. While cameras were deployed in the field for longer in 2024, this did not result in a significantly larger amount of detection events. (c-e) Distributions of Activity, Species, and Action labels for both seasons. We indicate the total number of videos containing a given attribute class on top of each bar. The distributions for Species and Action labels are shown in log-scale for visibility.}  
    \label{app-fig:mammals-s2}
\end{figure}

We compare both seasons of data constituting MammAlps-S2, after label refinement and annotation (\Cref{app-fig:mammals-s2}).
The number of video hours acquired per site in 2024 follows the same general trend as in 2023. The hardware of C6 in S3 failed early and captured a single event only. The distribution of events per species in 2024 is slightly less skewed than in 2023, leading to an increased diversity for rare species. On the contrary, behavioral expressions were more diverse in 2023. This is partly explained by the fact that the 2023 data acquisition period was during the red deer rutting season, which led to more detections and diverse behaviors that were not seen in 2024, for which the data acquisition period was earlier in the summer.

Overall 101 false positive videos were added to be representative of real fieldwork datasets. In practice, false positive triggers can represent a much larger proportion of detected events, but replicating this ratio would lead to unnecessarily high computations. 

\subsection{Training and evaluation splits}
\label{app-sec:video-split}
\begin{figure}[ht!]
    \centering
    \includegraphics[width=1.0\linewidth]{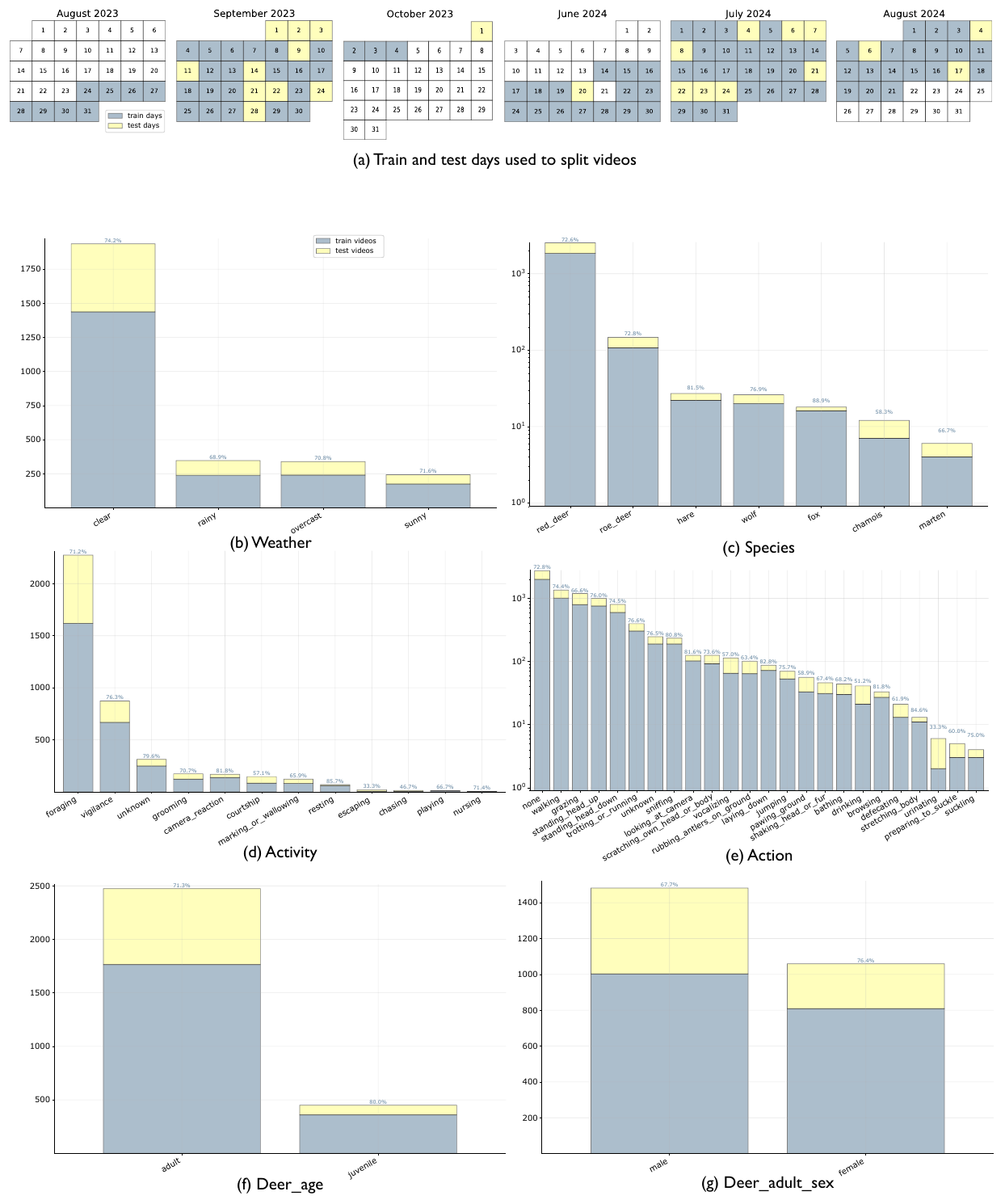}
    \caption{\textbf{Attribute distributions at the video-level for the train and test splits.} (a) Attribution of the train and test days along the acquisition periods. (b-g)~Distribution of labels per split for each attribute category. We indicate the percentage of videos containing a given label in the train set on top of each bar. The distributions of videos per Species and Action labels are shown in log-scale for visibility.}
    \label{app-fig:attributes_per_split}
\end{figure}

Camera-trap events were randomly split at the day-level, while manually ensuring a relatively similar label distribution across splits. We represent training and testing days, along with the distribution of classes for each attribute per split in~\cref{app-fig:attributes_per_split}. The \textsl{none} action is the default value for the optional second action and is not used to train SALMA.

\subsection{Text-to-video associations}
\label{app-sec:benchmark}

\begin{figure}
    \centering
    \includegraphics[width=0.95\linewidth]{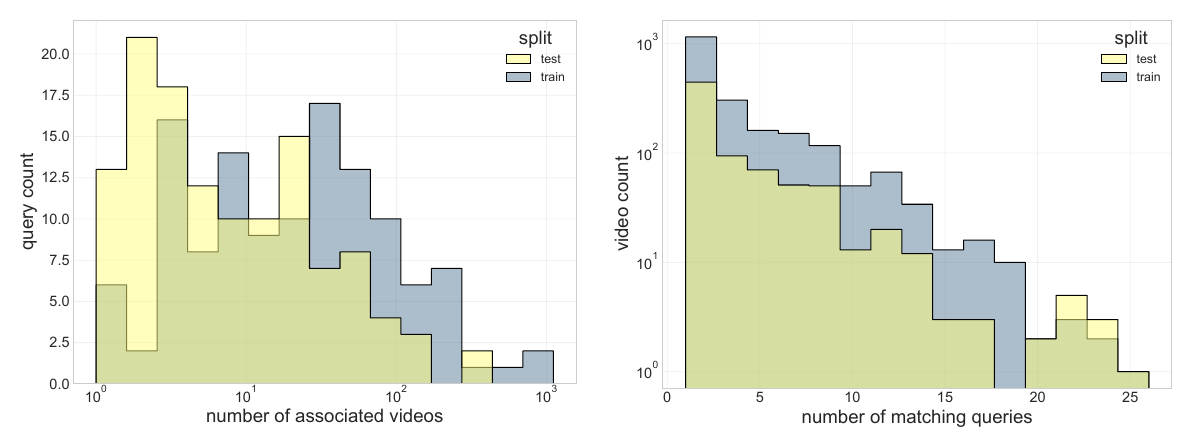}
        \caption{\textbf{Distributions of text-to-video associations.} (Left) Histogram of the number of videos matching a given query; (right) histogram of the number of queries matching a given video. Null entries \textendash empty queries or videos not matched to any element \textendash are not shown in this visualization.} 
    \label{app-fig:distributions_per_split}
\end{figure}

We list the proposed queries for the Prompting-MammAlps benchmark in \cref{app-tab:queries}, and indicate their respective number of associated videos for both the train and test splits. We show the distribution of the number of videos per query, and of the number of queries matching a given video for each split in \cref{app-fig:distributions_per_split}.

\clearpage
\renewcommand{\arraystretch}{1.5}
\tiny
\begin{longtable}{l p{0.6\textwidth} c c}
\caption{\textbf{Prompting-MammAlps queries and associated number of videos.} The train and test (\ie candidate) sets contain 2090 and 775 videos, respectively.}
\label{app-tab:queries} \\
\toprule
\textbf{id} &
  \textbf{Query} &
  \textbf{\# Train videos} &
  \textbf{\# Test videos} \\
\midrule
\endfirsthead
\multicolumn{4}{c}%
{{\bfseries Table \thetable\ continued from previous page}} \\
\toprule
  \textbf{id} &
  \textbf{Query} &
  \textbf{\# Train videos} &
  \textbf{\# Test videos} \\
\midrule
\endhead
\textbf{0:} &
  \query{An animal engaged in any activity other than foraging.} &
  1110 &
  329 \\
\textbf{1:} &
  \query{An animal that is neither a red deer nor a roe deer.} &
  69 &
  20 \\
\textbf{2:} &
  \query{An animal running.} &
  304 &
  93 \\
\textbf{3:} &
  \query{An animal bathing.} &
  28 &
  14 \\
\textbf{4:} &
  \query{A roe deer grazing.} &
  14 &
  10 \\
\textbf{5:} &
  \query{An animal browsing.} &
  27 &
  6 \\
\textbf{6:} &
  \query{An adult roe deer sniffing.} &
  7 &
  5 \\
\textbf{7:} &
  \query{A juvenile red deer scratching its body.} &
  41 &
  10 \\
\textbf{8:} &
  \query{A juvenile roe deer preparing to suckle.} &
  \textbf{0} &
  \textbf{0} \\
\textbf{9:} &
  \query{A chamois trotting.} &
  4 &
  2 \\
\textbf{10:} &
  \query{An adult male roe deer jumping.} &
  4 &
  2 \\
\textbf{11:} &
  \query{An adult male red deer rubbing its antlers on the ground.} &
  62 &
  34 \\
\textbf{12:} &
  \query{A fox sniffing.} &
  1 &
  \textbf{0} \\
\textbf{13:} &
  \query{A fox chasing prey.} &
  \textbf{0} &
  \textbf{0} \\
\textbf{14:} &
  \query{A wolf chasing prey.} &
  4 &
  4 \\
\textbf{15:} &
  \query{A hare.} &
  22 &
  5 \\
\textbf{16:} &
  \query{A marten.} &
  4 &
  2 \\
\textbf{17:} &
  \query{A juvenile red deer playing.} &
  10 &
  4 \\
\textbf{18:} &
  \query{An adult red deer browsing.} &
  14 &
  1 \\
\textbf{19:} &
  \query{A hare foraging.} &
  15 &
  4 \\
\textbf{20:} &
  \query{A chamois.} &
  7 &
  5 \\
\textbf{21:} &
  \query{A chamois being vigilant.} &
  5 &
  2 \\
\textbf{22:} &
  \query{A fox foraging.} &
  16 &
  2 \\
\textbf{23:} &
  \query{A wolf foraging.} &
  14 &
  2 \\
\textbf{24:} &
  \query{A juvenile red deer shaking its head or body.} &
  5 &
  2 \\
\textbf{25:} &
  \query{A juvenile roe deer.} &
  3 &
  \textbf{0} \\
\textbf{26:} &
  \query{A roe deer trotting while foraging.} &
  29 &
  7 \\
\textbf{27:} &
  \query{A mountain hare being vigilant.} &
  4 &
  3 \\
\textbf{28:} &
  \query{A mountain hare grazing.} &
  1 &
  1 \\
\textbf{29:} &
  \query{An animal stretching its body.} &
  11 &
  2 \\
\textbf{30:} &
  \query{An animal bathing while grooming.} &
  17 &
  8 \\
\textbf{31:} &
  \query{An animal resting.} &
  60 &
  10 \\
\textbf{32:} &
  \query{A red deer browsing.} &
  17 &
  1 \\
\textbf{33:} &
  \query{An animal running while foraging.} &
  142 &
  37 \\
\textbf{34:} &
  \query{A video of an animal drinking.} &
  21 &
  20 \\
\textbf{35:} &
  \query{A video of an animal lying down while resting.} &
  58 &
  10 \\
\textbf{36:} &
  \query{A red deer resting in rainy weather.} &
  19 &
  2 \\
\textbf{37:} &
  \query{A red deer resting in clear weather.} &
  36 &
  4 \\
\textbf{38:} &
  \query{Rainy weather.} &
  239 &
  108 \\
\textbf{39:} &
  \query{A juvenile deer in clear or sunny weather.} &
  257 &
  54 \\
\textbf{40:} &
  \query{A juvenile deer in rainy weather.} &
  47 &
  8 \\
\textbf{41:} &
  \query{An animal participating in courtship.} &
  84 &
  63 \\
\textbf{42:} &
  \query{An adult red deer standing with its head up while participating in courtship.} &
  47 &
  41 \\
\textbf{43:} &
  \query{An adult red deer vocalizing while participating in courtship.} &
  62 &
  49 \\
\textbf{44:} &
  \query{A video of two adult red deer vocalizing while participating in courtship.} &
  1 &
  \textbf{0} \\
\textbf{45:} &
  \query{An adult red deer lying down while participating in courtship.} &
  7 &
  1 \\
\textbf{46:} &
  \query{An adult male red deer running while participating in courtship.} &
  9 &
  6 \\
\textbf{47:} &
  \query{An adult female red deer running while participating in courtship.} &
  3 &
  13 \\
\textbf{48:} &
  \query{A video of one adult male and one adult female red deer running while participating in courtship.} &
  3 &
  4 \\
\textbf{49:} &
  \query{An adult male roe deer running while participating in courtship.} &
  15 &
  3 \\
\textbf{50:} &
  \query{An adult female roe deer running while participating in courtship.} &
  12 &
  3 \\
\textbf{51:} &
  \query{A video of one adult male and one adult female roe deer running while participating in courtship.} &
  11 &
  3 \\
\textbf{52:} &
  \query{An adult male red deer pawing the ground while wallowing.} &
  28 &
  22 \\
\textbf{53:} &
  \query{An adult male red deer pawing the ground and rubbing its antlers on the ground while wallowing.} &
  20 &
  18 \\
\textbf{54:} &
  \query{An adult male red deer bathing while wallowing.} &
  7 &
  1 \\
\textbf{55:} &
  \query{An adult male red deer urinating while wallowing.} &
  1 &
  2 \\
\textbf{56:} &
  \query{An adult male red deer being vigilant after vocalizing.} &
  29 &
  25 \\
\textbf{57:} &
  \query{A red deer vocalizing in rainy weather.} &
  \textbf{0} &
  3 \\
\textbf{58:} &
  \query{A roe deer participating in courtship in clear weather.} &
  5 &
  3 \\
\textbf{59:} &
  \query{A roe deer participating in courtship in rainy weather.} &
  8 &
  0 \\
\textbf{60:} &
  \query{A red deer participating in courtship in rainy weather.} &
  \textbf{0} &
  4 \\
\textbf{61:} &
  \query{A video of two or more animals.} &
  265 &
  97 \\
\textbf{62:} &
  \query{A video of three or more red deer.} &
  47 &
  23 \\
\textbf{63:} &
  \query{A video of two red deer.} &
  193 &
  60 \\
\textbf{64:} &
  \query{A video of two or more red deer escaping.} &
  6 &
  9 \\
\textbf{65:} &
  \query{A juvenile red deer walking while being vigilant.} &
  52 &
  5 \\
\textbf{66:} &
  \query{A video of two adult female red deer walking while foraging.} &
  23 &
  11 \\
\textbf{67:} &
  \query{A video of three or more adult male red deer walking while foraging.} &
  \textbf{0} &
  \textbf{0} \\
\textbf{68:} &
  \query{A video of two or more wolves.} &
  \textbf{0} &
  1 \\
\textbf{69:} &
  \query{An adult female red deer foraging and a juvenile red deer foraging.} &
  124 &
  56 \\
\textbf{70:} &
  \query{A video showing individuals from at least two different species.} &
  7 &
  3 \\
\textbf{71:} &
  \query{A juvenile red deer nursing.} &
  5 &
  2 \\
\textbf{72:} &
  \query{A juvenile red deer suckling.} &
  3 &
  1 \\
\textbf{73:} &
  \query{A juvenile roe deer suckling.} &
  \textbf{0} &
  \textbf{0} \\
\textbf{74:} &
  \query{An animal escaping from another animal of the same species that is chasing it.} &
  2 &
  4 \\
\textbf{75:} &
  \query{An animal escaping from another animal of a different species that is chasing it.} &
  4 &
  2 \\
\textbf{76:} &
  \query{A video of two or more juvenile red deer.} &
  24 &
  8 \\
\textbf{77:} &
  \query{Two or more red deer foraging, with at least one also being vigilant at some point.} &
  106 &
  26 \\
\textbf{78:} &
  \query{A juvenile red deer doing the exact same activities as an adult female red deer.} &
  69 &
  34 \\
\textbf{79:} &
  \query{A juvenile red deer doing the exact same actions as an adult female red deer.} &
  31 &
  23 \\
\textbf{80:} &
  \query{A juvenile red deer doing at least one different action than an adult female red deer.} &
  145 &
  45 \\
\textbf{81:} &
  \query{An adult female red deer nursing.} &
  2 &
  2 \\
\textbf{82:} &
  \query{A wolf chasing prey in clear or sunny weather.} &
  4 &
  2 \\
\textbf{83:} &
  \query{A wolf chasing prey in overcast or rainy weather.} &
  \textbf{0} &
  2 \\
\textbf{84:} &
  \query{A fox chasing prey in clear weather.} &
  \textbf{0} &
  \textbf{0} \\
\textbf{85:} &
  \query{A roe deer foraging in sunny weather.} &
  23 &
  3 \\
\textbf{86:} &
  \query{A red deer foraging in sunny weather.} &
  51 &
  33 \\
\textbf{87:} &
  \query{A roe deer foraging in rainy weather.} &
  7 &
  3 \\
\textbf{88:} &
  \query{A red deer foraging in rainy weather.} &
  186 &
  79 \\
\textbf{89:} &
  \query{An adult female red deer escaping in sunny weather.} &
  0 &
  0 \\
\textbf{90:} &
  \query{An animal being vigilant while the weather is rainy or overcast.} &
  211 &
  90 \\
\textbf{91:} &
  \query{An animal being vigilant while the weather is clear or sunny.} &
  456 &
  117 \\
\textbf{92:} &
  \query{An animal reacting to a camera.} &
  139 &
  31 \\
\textbf{93:} &
  \query{Two or more animals reacting to a camera.} &
  8 &
  1 \\
\textbf{94:} &
  \query{An adult roe deer reacting to a camera.} &
  4 &
  1 \\
\textbf{95:} &
  \query{A video of three or more red deer reacting to a camera.} &
  \textbf{0} &
  1 \\
\textbf{96:} &
  \query{An adult red deer reacting to a camera.} &
  103 &
  26 \\
\textbf{97:} &
  \query{A juvenile deer reacting to a camera.} &
  36 &
  4 \\
\textbf{98:} &
  \query{An adult male red deer reacting to a camera.} &
  44 &
  11 \\
\textbf{99:} &
  \query{A video of an individual sniffing while reacting to a camera.} &
  33 &
  5 \\
\textbf{100:} &
  \query{An adult female red deer reacting to a camera.} &
  59 &
  15 \\
\textbf{101:} &
  \query{An animal running while reacting to a camera.} &
  34 &
  11 \\
\textbf{102:} &
  \query{An animal looking at a camera.} &
  102 &
  23 \\
\textbf{103:} &
  \query{A fox reacting to a camera.} &
  \textbf{0} &
  \textbf{0} \\
\textbf{104:} &
  \query{A wolf reacting to a camera.} &
  3 &
  \textbf{0} \\
\textbf{105:} &
  \query{An animal reacting to a camera and then foraging.} &
  47 &
  6 \\
\textbf{106:} &
  \query{An animal foraging, then reacting to a camera, and then returning to foraging.} &
  23 &
  5 \\
\textbf{107:} &
  \query{An animal reacting to a camera and then running away.} &
  35 &
  11 \\
\textbf{108:} &
  \query{An animal foraging, then reacting to a camera, and then running away.} &
  8 &
  4 \\
\textbf{109:} &
  \query{An animal reacting to a camera in rainy or overcast weather.} &
  41 &
  5 \\
\textbf{110:} &
  \query{An animal reacting to a camera in clear or sunny weather.} &
  98 &
  26 \\
\textbf{111:} &
  \query{An adult male red deer doing the same sequence of actions while wallowing as any individual of \vidtag{S3_C2_E769_V0436}.} &
  \textbf{0} &
  1 \\
\textbf{112:} &
  \query{An adult male red deer doing the same sequence of actions while wallowing as any individual of \vidtag{S3_C3_E524_V0327}.} &
  1 &
  2 \\
\textbf{113:} &
  \query{An adult male red deer doing the same unique actions while wallowing as any individual of \vidtag{S3_C3_E524_V0327}.} &
  7 &
  5 \\
\textbf{114:} &
  \query{An adult male red deer doing the same unique actions while wallowing as any individual of \vidtag{S3_C2_E524_V0087}.} &
  6 &
  6 \\
\textbf{115:} &
  \query{An adult male red deer performing at least one different action while wallowing than with any individual of \vidtag{S3_C3_E524_V0327}.} &
  27 &
  12 \\
\textbf{116:} &
  \query{A juvenile red deer performing at least one action in common while nursing with any individual in \vidtag{S1_C6_F394_V0310}.} &
  5 &
  2 \\
\textbf{117:} &
  \query{A juvenile red deer performing at least one action in common while nursing with any individual in \vidtag{S1_C6_F404_V0330}.} &
  5 &
  2 \\
\textbf{118:} &
  \query{A juvenile red deer performing at least one different action while nursing than with any individual in \vidtag{S1_C6_F404_V0330}.} &
  4 &
  1 \\
\textbf{119:} &
  \query{An individual doing the same sequence of actions while reacting to a camera as any individual of \vidtag{S1_C4_F173_V0137}.} &
  7 &
  2 \\
\textbf{120:} &
  \query{An individual doing the same sequence of actions while reacting to a camera as any individual of \vidtag{S1_C3_E144_V0060}.} &
  78 &
  16 \\
\textbf{121:} &
  \query{An individual doing the same unique actions while reacting to a camera as any individual of \vidtag{S1_C4_F173_V0137}.} &
  3 &
  2 \\
\textbf{122:} &
  \query{An individual doing the same unique actions while reacting to a camera as any individual of \vidtag{S1_C3_E144_V0060}.} &
  37 &
  9 \\
\textbf{123:} &
  \query{An individual doing the same sequence of actions as the individual of \vidtag{S2_C1_E323_V0074}.} &
  1 &
  1 \\
\textbf{124:} &
  \query{An individual doing at least one different action while reacting to a camera than with any individual of \vidtag{S1_C4_F173_V0137}.} &
  114 &
  26 \\
\textbf{125:} &
  \query{An individual doing at least one different action while reacting to a camera than with any individual of \vidtag{S1_C3_E144_V0060}.} &
  102 &
  21 \\
\textbf{126:} &
  \query{An animal being vigilant in a different weather condition from that in \vidtag{S2_C2_F536_V0066}.} &
  250 &
  120 \\
\textbf{127:} &
  \query{A wolf chasing prey in a different weather condition from that in \vidtag{S2_C1_F573_V0093}.} &
  4 &
  2 \\
\textbf{128:} &
  \query{An individual sharing at least one activity with any individual from \vidtag{S2_C1_F573_V0093}.} &
  8 &
  16 \\
\textbf{129:} &
  \query{An individual sharing at least one activity with any individual from \vidtag{S3_C2_E670_V0159} but in a different weather condition.} &
  42 &
  53 \\
\textbf{130:} &
  \query{An individual sharing at least one activity with any individual of the same species from \vidtag{S1_C4_F136_V0115} but in a different weather condition.} &
  57 &
  28 \\
\textbf{131:} &
  \query{An individual sharing at least one activity with any individual from \vidtag{S1_C1_E66_V0152} but of a different species.} &
  161 &
  49 \\
\textbf{132:} &
  \query{An individual performing the same activities as any individual from \vidtag{S2_C2_E351_V0099} but in a different weather condition.} &
  30 &
  18 \\
\textbf{133:} &
  \query{A single adult red deer foraging only.} &
  810 &
  373 \\
\textbf{134:} &
  \query{An empty video.} &
  78 &
  23 \\
\bottomrule
\end{longtable}
\normalsize

\subsection{SALMA}
\label{app-sec:salma}

\subsubsection{Details of SALMA's architecture.}

SALMA is a transformer encoder-decoder architecture inspired by Carion et al.~\cite{carion2020end, zeng2022motr}. The encoder is the same vision transformer encoder as used in VideoMAE~\cite{tong2022videomae}. Input frames were first transformed into crops using a random cropping with scaling strategy, and further augmented with horizontal flipping, and color transformations. The input to the transformer encoder consisted of 16 frame crops of size (224, 224) that were tokenized using a temporal patch size of two frames, and a spatial patch size of 16 pixels, yielding 1568 tokens of embedding dimension 1024. A learnable positional encoding was added to each token of the input sequence. The encoder contained a sequence of 24 transformer blocks each containing 16 heads. The transformer decoder was used to progressively decode 20 object queries of embedding dimension 768. It contained four multi-head transformer layers, each consisting of self-attention followed by cross-attention with the encoded sequence of tokens, and layer normalization. We used eight heads for the transformer decoder. The output of each decoder step was given as input to six different multi-layer perceptron (MLP) heads (one per attribute category, and one for the bounding box coordinates and track activation score prediction), each comprising of one hidden layer containing 192 neurons. A final MLP head projected the averaged sequence of encoded tokens to four output neurons representing the weather conditions. 
ReLU activation was used in the encoder, the decoder, and in the MLP heads.

\subsubsection{Training SALMA.}
After each forward pass, we computed 11 loss terms: one cross-entropy (CE) loss for each attribute category (using binary cross-entropy (BCE) for actions), a gIoU and a $\ell_1$-loss for the bounding box coordinates, a BCE loss on the object class using the track activation score prediction, and a contrastive loss on the object queries following~\cite{manasyan2025temporally}.
We back-propagated the weighted loss values using the weighted Adam optimizer \cite{kingma2014adam} with $\beta_1=0.90$ and $\beta_2=0.95$. We applied dropout to 10\% of all weights and used a weight decay value of $10^{-3}$.

In practice, we split the training of SALMA in two curriculum learning steps, where for the first 500 epochs, only the parameters of the encoder, the decoder, and the MLP head used for bounding box coordinate and activation score predictions were updated. For the following 500 epochs, all model parameters were optimized. 
To improve the robustness to variations in temporal striding of the input sequence at inference, to variations in the speed at which a behavior might be expressed, and to better model long-term dynamics (\ie occlusions), we sampled a random temporal stride value between one and 30 frames, leading to input sequences spanning 0.5 seconds to 16 seconds. The start of the input sequence was also randomly sampled within the video, when possible.
To compensate for the strong class imbalance, we sampled videos using a weight corresponding to their averaged inverse class frequencies among class attributes, which we further divided by three whenever a video contains more than one individual. 
Training was performed on eight V100 NVIDIA GPUs, using a total batch size of 64 frame sequences.

\subsubsection{Inference.}
Video frames have a resolution of $1920 \times 1080$ pixels, while SALMA was trained on square image crops. Each frame was therefore split into two $1080 \times 1080$ square crops covering the left and right parts of the image, with an overlapping region at the center. Each crop was then down-sampled to $224 \times 224$ pixels and inference was performed on both views independently. As a result, objects appearing in the overlapping region may be detected twice and assigned a different track identifier, depending on which object query became activated. To reconcile these predictions, we matched detections from the left and right views within the overlapping region using a class-agnostic Hungarian matcher \cite{kuhn1955hungarian}. The matching cost was defined from the bounding-box gIoU and $\ell_1$ coordinate distance. Matched detections were then assigned to a common track identifier, allowing for the corresponding tracks to be merged across the two views. After merging, we applied few post-processing steps to improve track consistency. First, detections localized only within the overlapping region were removed, as these are typically redundant artifacts produced by the dual inference. We then discarded tracks spanning fewer than five frames. Our cross-view merging strategy can lead to two detections with the same track identifier. In these cases, we removed the track with the lowest confidence score. Finally, we performed a track-level non-maxima suppression (NMS) step: for each pair of tracks with more than $70\%$ of temporally overlapping boxes and with a gIoU greater than $0.8$, we computed the sum of detection confidence scores and removed the track with the lowest total score.

\subsubsection{Detailed F1-score per attribute class.} We detail the F1-scores per class attribute in \cref{app-tab:per_class_f1}.

\renewcommand{\arraystretch}{0.6}
\scriptsize
\setlength\tabcolsep{3pt}
\begin{longtable}{p{0.25\textwidth} p{0.30\textwidth} p{0.1\textwidth} p{0.25\textwidth}}
\caption{\textbf{Detailed per-class F1-scores for each classification attribute.}}
\label{app-tab:per_class_f1}\\
\toprule
\textbf{Category} & \textbf{Class} & \textbf{F1} & \textbf{Support (detections)}\\ 
\midrule
\endfirsthead

\multicolumn{4}{c}{{\bfseries Table \thetable\ continued from previous page}} \\
\toprule
\textbf{Category} & \textbf{Class} & \textbf{F1} & \textbf{Support (detections)}\\
\midrule
\endhead

\multicolumn{3}{l}{\textbf{Species}} \\
 & Chamois & 0.00 & 6989\\
 & Fox & 0.64  & 1216\\
 & Hare & 0.60 & 2044\\
 & Marten & 0.00 & 608\\
 & Red Deer & 0.91 & 436886\\
 & Roe Deer & 0.49 & 28133\\
 & Wolf & 0.69 & 564\\ 
 & \textit{Micro avg.} & 0.88  & 476440\\
 & \textit{Macro avg.} & 0.48 & - \\
\midrule
\multicolumn{3}{l}{\textbf{Activity}} \\
 & Camera Reaction & 0.20 & 5044\\
 & Chasing & 0.42 & 560\\
 & Courtship & 0.50 & 20069\\
 & Escaping & 0.00 & 1746\\
 & Foraging & 0.83 & 318654\\
 & Grooming & 0.30 & 11898\\
 & Marking or Wallowing & 0.74 & 17230\\
 & Nursing & 0.00 & 743\\
 & Playing & 0.17 & 1674\\
 & Resting & 0.76 & 5236\\
 & Unknown & 0.08 & 10849\\
 & Vigilance & 0.72 & 82737\\
 & \textit{Micro avg.} & 0.77 & 476440\\
 & \textit{Macro avg.} & 0.40 & -\\
\midrule

\multicolumn{3}{l}{\textbf{Action}} \\
 & Bathing & 0.67 & 9899\\
 & Browsing & 0.00 & 1989\\
 & Defecating & 0.00 & 2219\\
 & Drinking & 0.62 & 8869\\
 & Grazing & 0.78 & 185292\\
 & Jumping & 0.16 & 1321\\
 & Laying Down & 0.50 & 8818\\
 & Looking at Camera & 0.09 & 3194\\
 & Pawing Ground & 0.35 & 5204\\
 & Preparing to Suckle & 0.00 & 136\\
 & Rubbing Antlers on Ground & 0.58 & 9215\\
 & Scratching own Head or Body & 0.09 & 5984\\
 & Shaking Head or Fur & 0.15 & 963\\
 & Sniffing & 0.16 & 5450\\
 & Standing Head Down & 0.29 & 30507\\
 & Standing Head Up & 0.74 & 102628\\
 & Stretching Body & 0.00 & 343\\
 & Suckling & 0.00 & 207\\
 & Trotting or Running & 0.37 & 10975\\
 & Unknown & 0.13 & 12591\\
 & Urinating & 0.00 & 2105\\
\ & Vocalizing & 0.45 & 10390\\
 & Walking & 0.57 & 83864\\
 & \textit{Micro avg.} & 0.59 & 502163\\
 & \textit{Macro avg.} & 0.29 & -\\
\midrule

\multicolumn{3}{l}{\textbf{Deer Age}} \\
 & Adult & 0.89 & 411636 \\
 & Juvenile & 0.51 & 53383\\
 & \textit{Micro avg.} & 0.85 & 465019\\
 & \textit{Macro avg.} & 0.70 & -\\
\midrule
\newpage
\multicolumn{3}{l}{\textbf{Deer Adult Sex}} \\
 & Female & 0.64 & 150607\\
 & Male & 0.86 & 261029\\
 & \textit{Micro avg.} & 0.77 & 411636\\
 & \textit{Macro avg.} & 0.75 & -\\
 \vspace{6pt} \\
\toprule
\textbf{Category} & \textbf{Class} & \textbf{F1} & \textbf{Support (videos)}\\
\midrule
\multicolumn{3}{l}{\textbf{Weather}} \\
 & Clear & 0.91 & 499\\
 & Overcast & 0.58 & 99\\
 & Rainy & 0.68 & 108\\
 & Sunny & 0.88 & 69\\
 & \textit{Micro avg.} & 0.84 & 775\\
 & \textit{Macro avg.} & 0.76 & -\\

\bottomrule
\end{longtable}
\normalsize




\subsection{Evaluation of vision-language models (VLMs)}
\label{app-sec:sota}
For CLIP4Clip~\cite{luo2021clip4clip}, we used the ViT-B/32 vision encoder backbone, sampled 12 frames at 1FPS and resized videos to 224x224 resolution. In SigLIP4Clip, we followed the same pipeline as for CLIP4Clip but replacing the visual backbone with SigLIP-SO400M\cite{zhai2023sigmoid}, and processing 32 frames at 384x384 resolution. For GRAM~\cite{cicchetti2024gramian}, we combined vision and audio modalities extracted from the videos, following default parameters. This was necessary as Gramian volume computation requires these three modalities (text, vision, and audio). With InternVideo2.0-1B~\cite{wang2022internvideo}, we sampled four frames at 224x224 resolution and computed embeddings using the 1B-parameters model. 
We used the code provided by~\cite{luo2021clip4clip} to fine-tune CLIP4Clip on the training data of MammAlps-S2. The query-video associations were transformed into 6852 text-video pairs. The best performance was obtained when fine-tuning the model for 6 epochs with a batch size of 64 sequences of 12 frames on a single NVIDIA v100 GPU.

Each method produces text and video embeddings separately, and computes a complete similarity matrix between the two modalities. CLIP4Clip, SigLIP4Clip, and InternVideo2.0-1B use cosine similarity as a similarity measure between modalities, while GRAM uses the Gramian volume between all three modalities (including audio). We returned the negative distance as a measure of similarity to maintain the same ranking as for the other three methods. 
Since our benchmark does not require to rank videos, similarity scores must be binarized to define the retrieved set of candidate videos. We calibrated a threshold for each query on the training split by maximizing the F1-score, and applied the resulting thresholds to the test set similarity scores. 
We evaluated the five models on a NVIDIA GeForce RTX 3090.

\subsection{Retrieval performance on individual queries}
\label{app-sec:per_query}

We report per-query F1-scores for the Agent+Oracle, Agent+SALMA, CLIP4-Clip~\cite{luo2021clip4clip}, and InternVideo2.0\cite{wang2022internvideo} models in \cref{app-tab:per_query_f1}. Empty queries \textendash those with no associated videos \textendash were handled by inverting both the ground-truth set and the set of predicted videos: the ground truth becomes the complete set of candidate videos and the prediction set becomes the set of non-retrieved (\ie filtered out) videos. This inversion ensures that empty queries contributed to the evaluation rather than being excluded (null support). Because there is no straightforward way to assess the performance of similarity-based VLMs on queries involving a reference video, we did not evaluate VLMs on these queries, and report a \enquote{n/a} value instead.

A F1-score $<1$ for the Agent+Oracle column indicates that the code generated by Qwen3-8B had an inaccurate retrieval logic. A F1-score $=0$ is a sign of either a wrong logic, or of a function that could not be executed (\eg syntax error).

\renewcommand{\arraystretch}{1.5}
\tiny
\begin{longtable}{p{0.5\textwidth} p{0.1\textwidth} p{0.1\textwidth} p{0.1\textwidth} p{0.1\textwidth}}
\caption{\textbf{Retrieval performance (F1-score) per query.}}
\label{app-tab:per_query_f1}\\
\toprule
\textbf{Query} &
  \textbf{Agent+ Oracle} &
  \textbf{Agent+ SALMA} &
  \textbf{CLIP4 Clip} &
  \textbf{Intern Video2.0} \\
\midrule
\endfirsthead
\multicolumn{5}{c}%
{{\bfseries Table \thetable\ continued from previous page}} \\
\toprule
\textbf{Query} &
  \textbf{Agent+ Oracle} &
  \textbf{Agent+ SALMA} &
  \textbf{CLIP4 Clip} &
  \textbf{Intern Video2.0} \\
\midrule
\endhead
\query{A juvenile roe deer preparing to suckle.} &
  1.0 &
  1.0 &
  1.0 &
  1.0 \\
\query{A fox sniffing.} &
  1.0 &
  1.0 &
  1.0 &
  1.0 \\
\query{A juvenile roe deer suckling.} &
  1.0 &
  1.0 &
  1.0 &
  1.0 \\
\query{An adult female red deer escaping in sunny weather.} &
  1.0 &
  1.0 &
  1.0 &
  1.0 \\
\query{A fox reacting to a camera.} &
  1.0 &
  1.0 &
  0.98 &
  0.98 \\
\query{A fox chasing prey.} &
  1.0 &
  1.0 &
  1.0 &
  1.0 \\
\query{A roe deer participating in courtship in rainy weather.} &
  1.0 &
  1.0 &
  1.0 &
  1.0 \\
\query{A fox chasing prey in clear weather.} &
  1.0 &
  1.0 &
  0.03 &
  0.15 \\
\query{A juvenile roe deer.} &
  1.0 &
  1.0 &
  0.98 &
  0.97 \\
\query{A wolf reacting to a camera.} &
  1.0 &
  1.0 &
  0.69 &
  0.57 \\
\query{An empty video.} &
  1.0 &
  0.93 &
  0.15 &
  0.15 \\
\query{A wolf chasing prey.} &
  1.0 &
  0.86 &
  0.04 &
  0.1 \\
\query{An animal resting.} &
  1.0 &
  0.86 &
  0.0 &
  0.4 \\
\query{A video of an animal lying down while resting.} &
  1.0 &
  0.86 &
  0.0 &
  0.0 \\
\query{A hare foraging.} &
  1.0 &
  0.8 &
  0.01 &
  0.01 \\
\query{A wolf chasing prey in a different weather condition from that in \vidtag{S2_C1_F573_V0093}.} &
  1.0 &
  0.8 &
  n/a &
  n/a \\
\query{An adult male red deer rubbing its antlers on the ground.} &
  1.0 &
  0.76 &
  0.21 &
  0.26 \\
\query{An animal being vigilant while the weather is rainy or overcast.} &
  1.0 &
  0.75 &
  0.58 &
  0.21 \\
\query{An animal bathing.} &
  1.0 &
  0.74 &
  0.0 &
  0.05 \\
\query{A hare.} &
  1.0 &
  0.73 &
  0.01 &
  0.0 \\
\query{An adult male red deer being vigilant after vocalizing.} &
  1.0 &
  0.7 &
  0.32 &
  0.61 \\
\query{An animal bathing while grooming.} &
  1.0 &
  0.7 &
  0.0 &
  0.02 \\
\query{Rainy weather.} &
  1.0 &
  0.68 &
  0.28 &
  0.36 \\
\query{An animal participating in courtship.} &
  1.0 &
  0.68 &
  0.35 &
  0.07 \\
\query{An animal running.} &
  1.0 &
  0.68 &
  0.22 &
  0.26 \\
\query{A red deer resting in clear weather.} &
  1.0 &
  0.67 &
  0.01 &
  0.01 \\
\query{A roe deer participating in courtship in clear weather.} &
  1.0 &
  0.67 &
  0.01 &
  0.0 \\
\query{A video of two or more wolves.} &
  1.0 &
  0.67 &
  0.0 &
  0.0 \\
\query{A red deer foraging in rainy weather.} &
  1.0 &
  0.64 &
  0.2 &
  0.18 \\
\query{An animal being vigilant while the weather is clear or sunny.} &
  1.0 &
  0.63 &
  0.26 &
  0.26 \\
\query{An adult male red deer pawing the ground while wallowing.} &
  1.0 &
  0.59 &
  0.19 &
  0.29 \\
\query{A video of two or more animals.} &
  1.0 &
  0.59 &
  0.22 &
  0.3 \\
\query{An adult male red deer pawing the ground and rubbing its antlers on the ground while wallowing.} &
  1.0 &
  0.59 &
  0.15 &
  0.25 \\
\query{A fox foraging.} &
  1.0 &
  0.57 &
  0.01 &
  0.0 \\
\query{A mountain hare being vigilant.} &
  1.0 &
  0.57 &
  0.01 &
  0.02 \\
\query{An adult female roe deer running while participating in courtship.} &
  1.0 &
  0.57 &
  0.0 &
  0.0 \\
\query{A video of an animal drinking.} &
  1.0 &
  0.49 &
  0.3 &
  0.08 \\
\query{A roe deer grazing.} &
  1.0 &
  0.47 &
  0.0 &
  0.03 \\
\query{A wolf foraging.} &
  1.0 &
  0.44 &
  0.02 &
  0.01 \\
\query{A juvenile deer in rainy weather.} &
  1.0 &
  0.44 &
  0.0 &
  0.02 \\
\query{An animal being vigilant in a different weather condition from that in \vidtag{S2_C2_F536_V0066}.} &
  1.0 &
  0.4 &
  n/a &
  n/a \\
\query{A video of two red deer.} &
  1.0 &
  0.38 &
  0.18 &
  0.25 \\
\query{A juvenile deer in clear or sunny weather.} &
  1.0 &
  0.38 &
  0.13 &
  0.13 \\
\query{An individual sharing at least one activity with any individual from \vidtag{S3_C2_E670_V0159} but in a different weather condition.} &
  1.0 &
  0.37 &
  n/a &
  n/a \\
\query{A juvenile red deer playing.} &
  1.0 &
  0.36 &
  0.1 &
  0.0 \\
\query{An animal looking at a camera.} &
  1.0 &
  0.36 &
  0.12 &
  0.06 \\
\query{An adult male red deer reacting to a camera.} &
  1.0 &
  0.34 &
  0.06 &
  0.1 \\
\query{A video of three or more red deer.} &
  1.0 &
  0.33 &
  0.19 &
  0.21 \\
\query{A video of an individual sniffing while reacting to a camera.} &
  1.0 &
  0.33 &
  0.0 &
  0.0 \\
\query{An adult male red deer performing at least one different action while wallowing than with any individual of \vidtag{S3_C3_E524_V0327}.} &
  1.0 &
  0.3 &
  n/a &
  n/a \\
\query{A juvenile red deer scratching its body.} &
  1.0 &
  0.26 &
  0.19 &
  0.02 \\
\query{A juvenile red deer shaking its head or body.} &
  1.0 &
  0.25 &
  0.07 &
  0.0 \\
\query{A roe deer foraging in sunny weather.} &
  1.0 &
  0.25 &
  0.07 &
  0.08 \\
\query{An animal reacting to a camera.} &
  1.0 &
  0.24 &
  0.1 &
  0.09 \\
\query{A video of two or more juvenile red deer.} &
  1.0 &
  0.21 &
  0.17 &
  0.33 \\
\query{An animal reacting to a camera in clear or sunny weather.} &
  1.0 &
  0.19 &
  0.1 &
  0.09 \\
\query{An individual doing at least one different action while reacting to a camera than with any individual of \vidtag{S1_C3_E144_V0060}.} &
  1.0 &
  0.18 &
  n/a &
  n/a \\
\query{A video showing individuals from at least two different species.} &
  1.0 &
  0.15 &
  0.0 &
  0.01 \\
\query{An animal reacting to a camera in rainy or overcast weather.} &
  1.0 &
  0.12 &
  0.08 &
  0.02 \\
\query{An animal browsing.} &
  1.0 &
  0.1 &
  0.04 &
  0.02 \\
\query{An animal running while reacting to a camera.} &
  1.0 &
  0.07 &
  0.06 &
  0.05 \\
\query{An individual sharing at least one activity with any individual of the same species from \vidtag{S1_C4_F136_V0115} but in a different weather condition.} &
  1.0 &
  0.05 &
  n/a &
  n/a \\
\query{An individual sharing at least one activity with any individual from \vidtag{S2_C1_F573_V0093}.} &
  1.0 &
  0.04 &
  n/a &
  n/a \\
\query{An individual doing the same sequence of actions as the individual of \vidtag{S2_C1_E323_V0074}.} &
  1.0 &
  0.03 &
  n/a &
  n/a \\
\query{An adult roe deer sniffing.} &
  1.0 &
  0.0 &
  0.06 &
  0.0 \\
\query{A chamois trotting.} &
  1.0 &
  0.0 &
  0.05 &
  0.0 \\
\query{An adult male roe deer jumping.} &
  1.0 &
  0.0 &
  0.03 &
  0.0 \\
\query{A marten.} &
  1.0 &
  0.0 &
  0.02 &
  0.01 \\
\query{An adult red deer browsing.} &
  1.0 &
  0.0 &
  0.01 &
  0.01 \\
\query{A chamois.} &
  1.0 &
  0.0 &
  0.0 &
  0.01 \\
\query{A chamois being vigilant.} &
  1.0 &
  0.0 &
  0.0 &
  0.0 \\
\query{A mountain hare grazing.} &
  1.0 &
  0.0 &
  0.0 &
  0.01 \\
\query{An animal stretching its body.} &
  1.0 &
  0.0 &
  0.0 &
  0.0 \\
\query{A red deer browsing.} &
  1.0 &
  0.0 &
  0.0 &
  0.11 \\
\query{A red deer resting in rainy weather.} &
  1.0 &
  0.0 &
  0.0 &
  0.02 \\
\query{A video of one adult male and one adult female roe deer running while participating in courtship.} &
  1.0 &
  0.0 &
  0.0 &
  0.02 \\
\query{An adult male red deer urinating while wallowing.} &
  1.0 &
  0.0 &
  0.0 &
  0.01 \\
\query{A red deer vocalizing in rainy weather.} &
  1.0 &
  0.0 &
  0.0 &
  0.0 \\
\query{A red deer participating in courtship in rainy weather.} &
  1.0 &
  0.0 &
  0.0 &
  0.0 \\
\query{A juvenile red deer nursing.} &
  1.0 &
  0.0 &
  0.0 &
  0.0 \\
\query{A juvenile red deer suckling.} &
  1.0 &
  0.0 &
  0.0 &
  0.0 \\
\query{An adult female red deer nursing.} &
  1.0 &
  0.0 &
  0.0 &
  0.0 \\
\query{A wolf chasing prey in overcast or rainy weather.} &
  1.0 &
  0.0 &
  0.0 &
  0.0 \\
\query{Two or more animals reacting to a camera.} &
  1.0 &
  0.0 &
  0.0 &
  0.0 \\
\query{An adult roe deer reacting to a camera.} &
  1.0 &
  0.0 &
  0.0 &
  0.0 \\
\query{A video of three or more red deer reacting to a camera.} &
  1.0 &
  0.0 &
  0.0 &
  0.0 \\
\query{An animal foraging, then reacting to a camera, and then returning to foraging.} &
  1.0 &
  0.0 &
  0.0 &
  0.0 \\
\query{An adult male red deer doing the same sequence of actions while wallowing as any individual of \vidtag{S3_C2_E769_V0436}.} &
  1.0 &
  0.0 &
  n/a &
  n/a \\
\query{A juvenile red deer performing at least one action in common while nursing with any individual in \vidtag{S1_C6_F394_V0310}.} &
  1.0 &
  0.0 &
  n/a &
  n/a \\
\query{A juvenile red deer performing at least one action in common while nursing with any individual in \vidtag{S1_C6_F404_V0330}.} &
  1.0 &
  0.0 &
  n/a &
  n/a \\
\query{A juvenile red deer performing at least one different action while nursing than with any individual in \vidtag{S1_C6_F404_V0330}.} &
  1.0 &
  0.0 &
  n/a &
  n/a \\
\query{A video of three or more adult male red deer walking while foraging.} &
  1.0 &
  0.99 &
  0.99 &
  0.99 \\
\query{A video of two adult red deer vocalizing while participating in courtship.} &
  0.99 &
  1.0 &
  0.57 &
  0.14 \\
\query{An adult female red deer foraging and a juvenile red deer foraging.} &
  0.97 &
  0.43 &
  0.19 &
  0.33 \\
\query{An animal engaged in any activity other than foraging.} &
  0.96 &
  0.78 &
  0.6 &
  0.6 \\
\query{An individual sharing at least one activity with any individual from \vidtag{S1_C1_E66_V0152} but of a different species.} &
  0.96 &
  0.56 &
  n/a &
  n/a \\
\query{A single adult red deer foraging only.} &
  0.96 &
  0.8 &
  0.64 &
  0.65 \\
\query{An adult red deer reacting to a camera.} &
  0.95 &
  0.2 &
  0.1 &
  0.14 \\
\query{A video of two or more red deer escaping.} &
  0.94 &
  0.0 &
  0.0 &
  0.15 \\
\query{An adult red deer standing with its head up while participating in courtship.} &
  0.93 &
  0.72 &
  0.43 &
  0.49 \\
\query{An individual doing at least one different action while reacting to a camera than with any individual of \vidtag{S1_C4_F173_V0137}.} &
  0.93 &
  0.16 &
  n/a &
  n/a \\
\query{An adult female red deer running while participating in courtship.} &
  0.93 &
  0.11 &
  0.09 &
  0.04 \\
\query{An adult female red deer reacting to a camera.} &
  0.88 &
  0.0 &
  0.03 &
  0.08 \\
\query{An animal reacting to a camera and then running away.} &
  0.88 &
  0.07 &
  0.05 &
  0.07 \\
\query{A red deer foraging in sunny weather.} &
  0.87 &
  0.65 &
  0.25 &
  0.34 \\
\query{An adult male roe deer running while participating in courtship.} &
  0.86 &
  0.0 &
  0.0 &
  0.0 \\
\query{Two or more red deer foraging, with at least one also being vigilant at some point.} &
  0.84 &
  0.31 &
  0.09 &
  0.17 \\
\query{An animal escaping from another animal of the same species that is chasing it.} &
  0.8 &
  0.0 &
  0.0 &
  0.0 \\
\query{An animal running while foraging.} &
  0.77 &
  0.37 &
  0.11 &
  0.09 \\
\query{A video of two adult female red deer walking while foraging.} &
  0.76 &
  0.17 &
  0.17 &
  0.16 \\
\query{A juvenile deer reacting to a camera.} &
  0.73 &
  0.12 &
  0.0 &
  0.0 \\
\query{An individual doing the same unique actions while reacting to a camera as any individual of \vidtag{S1_C3_E144_V0060}.} &
  0.72 &
  0.2 &
  n/a &
  n/a \\
\query{A juvenile red deer doing the exact same activities as an adult female red deer.} &
  0.69 &
  0.12 &
  0.02 &
  0.12 \\
\query{An adult male red deer doing the same sequence of actions while wallowing as any individual of \vidtag{S3_C3_E524_V0327}.} &
  0.67 &
  0.67 &
  n/a &
  n/a \\
\query{A juvenile red deer doing at least one different action than an adult female red deer.} &
  0.66 &
  0.29 &
  0.12 &
  0.21 \\
\query{An animal that is neither a red deer nor a roe deer.} &
  0.59 &
  0.4 &
  0.05 &
  0.05 \\
\query{A wolf chasing prey in clear or sunny weather.} &
  0.57 &
  0.5 &
  0.0 &
  0.0 \\
\query{An adult male red deer running while participating in courtship.} &
  0.57 &
  0.11 &
  0.08 &
  0.03 \\
\query{A juvenile red deer doing the exact same actions as an adult female red deer.} &
  0.55 &
  0.11 &
  0.06 &
  0.04 \\
\query{A video of one adult male and one adult female red deer running while participating in courtship.} &
  0.47 &
  0.0 &
  0.01 &
  0.03 \\
\query{An individual performing the same activities as any individual from \vidtag{S2_C2_E351_V0099} but in a different weather condition.} &
  0.47 &
  0.04 &
  n/a &
  n/a \\
\query{A juvenile red deer walking while being vigilant.} &
  0.45 &
  0.11 &
  0.07 &
  0.0 \\
\query{An individual doing the same sequence of actions while reacting to a camera as any individual of \vidtag{S1_C4_F173_V0137}.} &
  0.44 &
  0.07 &
  n/a &
  n/a \\
\query{An animal reacting to a camera and then foraging.} &
  0.43 &
  0.0 &
  0.13 &
  0.0 \\
\query{An adult male red deer doing the same unique actions while wallowing as any individual of \vidtag{S3_C3_E524_V0327}.} &
  0.33 &
  0.33 &
  n/a &
  n/a \\
\query{An adult red deer lying down while participating in courtship.} &
  0.33 &
  0.0 &
  0.0 &
  0.08 \\
\query{A roe deer trotting while foraging.} &
  0.32 &
  0.09 &
  0.06 &
  0.0 \\
\query{An adult male red deer bathing while wallowing.} &
  0.22 &
  0.12 &
  0.0 &
  0.04 \\
\query{A roe deer foraging in rainy weather.} &
  0.07 &
  0.06 &
  0.02 &
  0.01 \\
\query{An adult red deer vocalizing while participating in courtship.} &
  0.0 &
  0.0 &
  0.36 &
  0.48 \\
\query{An animal escaping from another animal of a different species that is chasing it.} &
  0.0 &
  0.0 &
  0.04 &
  0.04 \\
\query{An animal foraging, then reacting to a camera, and then running away.} &
  0.0 &
  0.0 &
  0.01 &
  0.08 \\
\query{An adult male red deer doing the same unique actions while wallowing as any individual of \vidtag{S3_C2_E524_V0087}.} &
  0.0 &
  0.0 &
  n/a &
  n/a \\
\query{An individual doing the same sequence of actions while reacting to a camera as any individual of \vidtag{S1_C3_E144_V0060}.} &
  0.0 &
  0.0 &
  n/a &
  n/a \\
\query{An individual doing the same unique actions while reacting to a camera as any individual of \vidtag{S1_C4_F173_V0137}.} &
  0.0 &
  0.0 &
  n/a &
  n/a \\
\bottomrule
\end{longtable}
\normalsize

\newpage
\subsection{Parsing functions library}
\label{app-sec:library}

We list the functions that the LLMs was encourage to use to process the candidate videos \texttt{.json} representations in \cref{app-tab:library}. These primitive functions were also used during the elaboration of the benchmark to associate the ground-truth video annotations from MammAlps-S2 to the queries. While the functions are specific to the proposed queries, they can easily be adapted or extended in practice to parse different information from the \texttt{.json} files. These primitive functions can be understood and trusted by the end-users, limit the risk of LLM hallucination during code generation, and ultimately improve interpretability of our TVR method.

\begin{table}[h]
\caption{\textbf{Library of parsing functions given to the LLMs.}}
\label{app-tab:library}
\setlength\tabcolsep{10pt}
\resizebox{\columnwidth}{!}{%
\begin{tabular}{lp{0.9\textwidth}}
\
\textbf{Function name}                            & \textbf{Description}                                                                          \\
\midrule
\texttt{get\_tracks\_from\_file\_id}              & Retrieves the individual tracks from a given \texttt{.json} file id.                           \\
\texttt{get\_weather\_condition\_from\_file\_id}  & Retrieves the weather condition from a given \texttt{.json} file id.                            \\
\texttt{check\_contains\_weather\_condition}      & Checks if the \texttt{file\_id} contains the specified weather condition                       \\
\texttt{get\_tracks\_from\_species}               & Retrieves tracks corresponding to a given species name.                               \\
\texttt{get\_tracks\_from\_action}                & Retrieves tracks corresponding to a given action name.                                \\
\texttt{get\_tracks\_from\_activity}              & Retrieves tracks corresponding to a given activity name.                              \\
\texttt{get\_deer\_tracks\_from\_age}             & Retrieves deer tracks corresponding to a given age group.                             \\
\texttt{get\_adult\_deer\_tracks\_from\_sex}      & Retrieves adult deer tracks corresponding to a given sex group.                       \\
\texttt{get\_nb\_tracks\_species\_in\_video}      & Returns the number of tracks for a given species in the given list of tracks.         \\
\texttt{get\_nb\_tracks\_action\_in\_video}       & Returns the number of tracks for a given action in the given list of tracks.          \\
\texttt{get\_nb\_tracks\_activity\_in\_video}     & Returns the number of tracks for a given activity in the given list of tracks.        \\
\texttt{get\_nb\_deer\_tracks\_age\_in\_video}    & Returns the number of deer tracks for a given age group in the given list of tracks.  \\
\texttt{get\_nb\_adult\_deer\_tracks\_sex\_in\_video} &
  Returns the number of adult deer tracks for a given sex group in the given list of tracks. \\
\texttt{get\_unique\_species\_from\_tracks}       & Returns the set of species present in all the individual tracks.                     \\
\texttt{get\_unique\_actions\_from\_tracks}       & Returns the set of actions present in all the individual tracks.                     \\
\texttt{get\_unique\_activities\_from\_tracks}    & Returns the set of activities present in all the individual tracks.                  \\
\texttt{get\_action\_sequences\_from\_tracks}     & Returns one sequence of actions for each individual track.                            \\
\texttt{check\_tracks\_contain\_species}          & Checks if any of the individual tracks contain the species of interest.              \\
\texttt{check\_tracks\_contain\_action}           & Checks if any of the individual tracks contain the action of interest.               \\
\texttt{check\_tracks\_contain\_activity}         & Checks if any of the individual tracks contain the activity of interest.             \\
\texttt{check\_tracks\_contain\_deer\_age}        & Checks if any of the individual tracks contain a deer of the given age group.        \\
\texttt{check\_tracks\_contain\_adult\_deer\_sex} & Checks if any of the individual tracks contain an adult deer of the given sex group. \\
\texttt{check\_track\_contains\_continuous\_sequence} &
  Checks if a given individual track contains a continuous series of behavior segments that contain attributes matching the sequence of interest. \\
\bottomrule
\end{tabular}%
}
\end{table}



\newpage
\subsection{Additional end-to-end retrieval results}
\label{app-sec:retrieval}

\begin{figure}[ht!]
    \centering
    \includegraphics[width=0.8\linewidth]{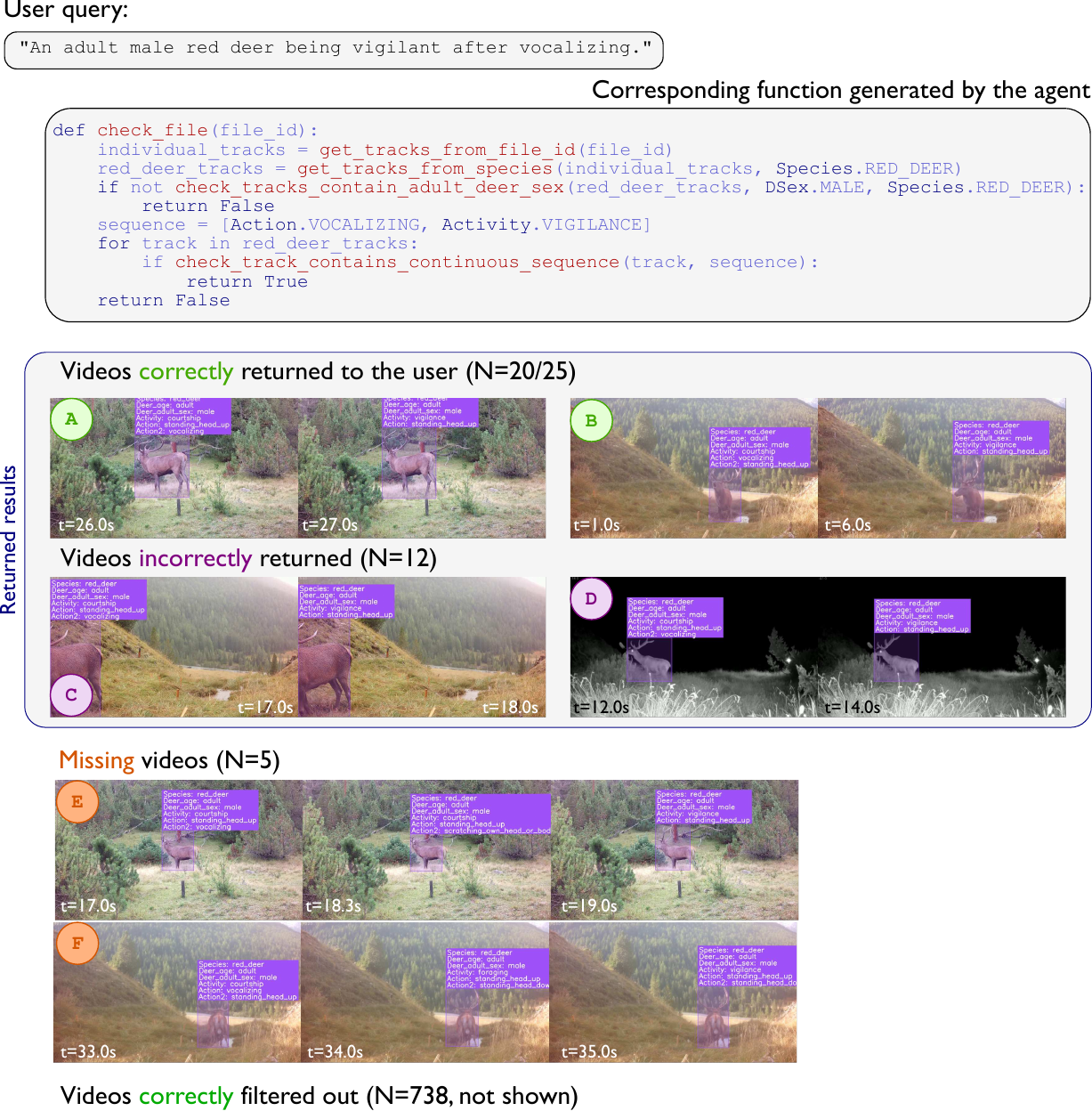}
    \caption{\textbf{Additional end-to-end retrieval result.} We used our final version of Qwen3-8B for prompt interpretation and code generation, and SALMA for frame-level predictions on the candidate videos.}
    \label{app-fig:decomposition_1}
\end{figure}

\begin{figure}[ht!]
    \centering
    \includegraphics[width=0.8\linewidth]{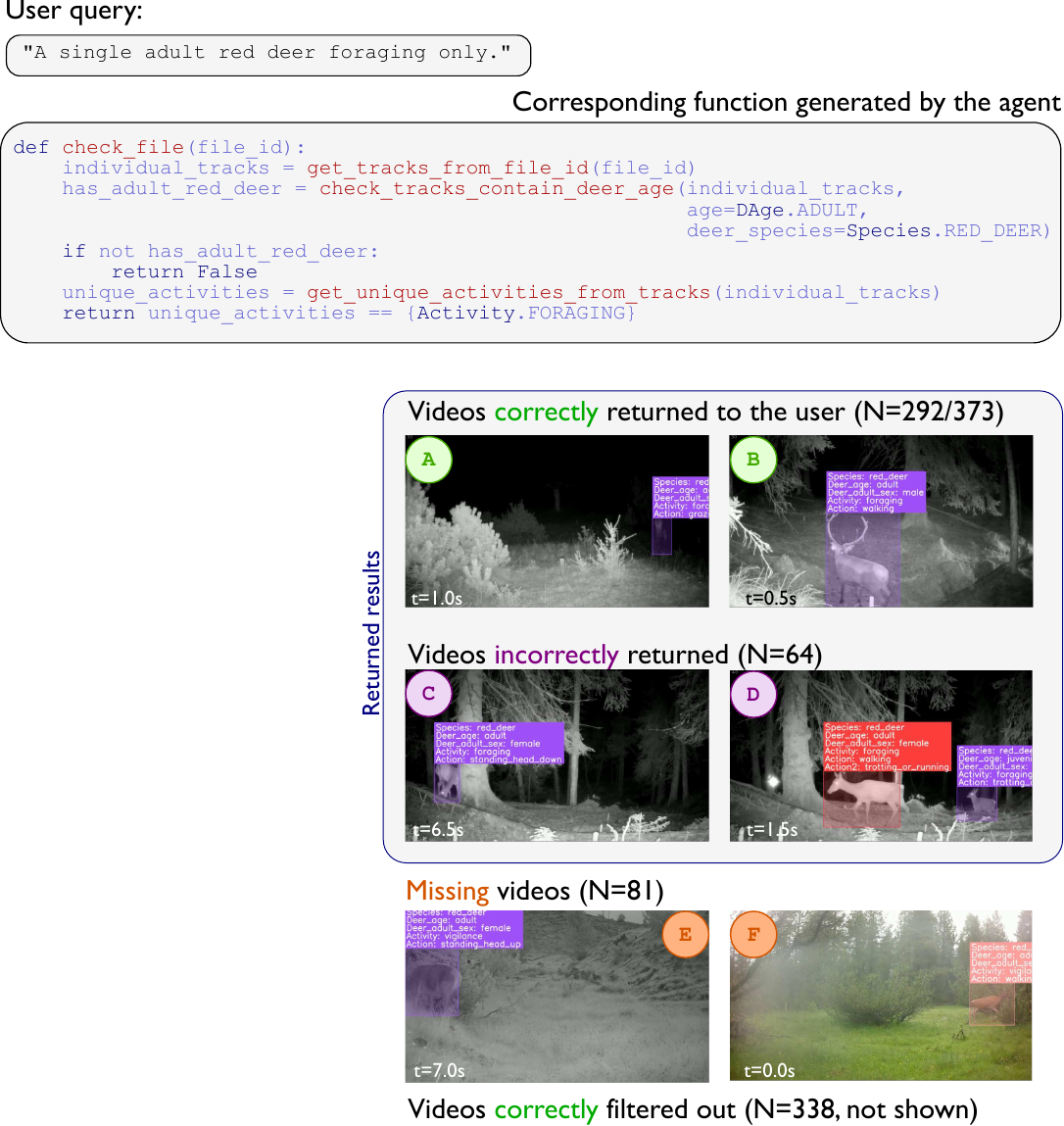}
    \caption{\textbf{Additional end-to-end retrieval result.} We used our final version of Qwen3-8B for prompt interpretation and code generation, and SALMA for frame-level predictions on the candidate videos.}
    \label{app-fig:decomposition_2}
\end{figure}

We show two additional end-to-end retrieval results in \cref{app-fig:decomposition_1} and \ref{app-fig:decomposition_2}. For the query \query{An adult male red deer being vigilant after vocalizing.}, we expected our approach to return videos where an adult male red deer performs the \textsl{vocalizing} action, directly followed by a \textsl{vigilant} activity. The code generated by Qwen3-8B follows this logic, and correctly addressed the distinction between the actions and activities. Out of the 25 videos returned to the user, 20 were correctly retrieved by our model (Agent+SALMA, A and B). We show two false positive results: in C, SALMA misclassified a segment as a \textsl{vigilant} behavior; in D, the animal was \textsl{standing head up}, but not \textsl{vocalizing}.
Five videos were missed: in E, \textsl{vocalizing} was followed by a misclassified \textsl{scratching own head or body} action; similarly in F, the vocalization was followed by a short misclassified \textsl{foraging} behavior. In both cases, the \textsl{vigilant} activity was well detected afterwards.

The query \query{A single adult red deer foraging only.} is designed to filter out very common events. The code generated by Qwen3-8B was slightly erroneous, as it did not check for the \enquote{single} individual condition. This explains why video D was incorrectly returned. Videos A and B were part of the 292 videos that were correctly retrieved. SALMA's prediction on video C did not contain any activity other than \textsl{foraging}, while the individual was actually \textsl{grooming} for few frames. Videos E and F are false negatives as both predicted results contain a wrong \textsl{vigilant} activity detection. The nuance between vigilance and foraging behaviors can be subtle when the animal is partially occluded (E) or walking (F).

\subsection{Video Captioning and Visual Question Answering}
\label{app-sec:extensions}

\begin{figure}[ht!]
    \centering
    \includegraphics[width=0.9\linewidth]{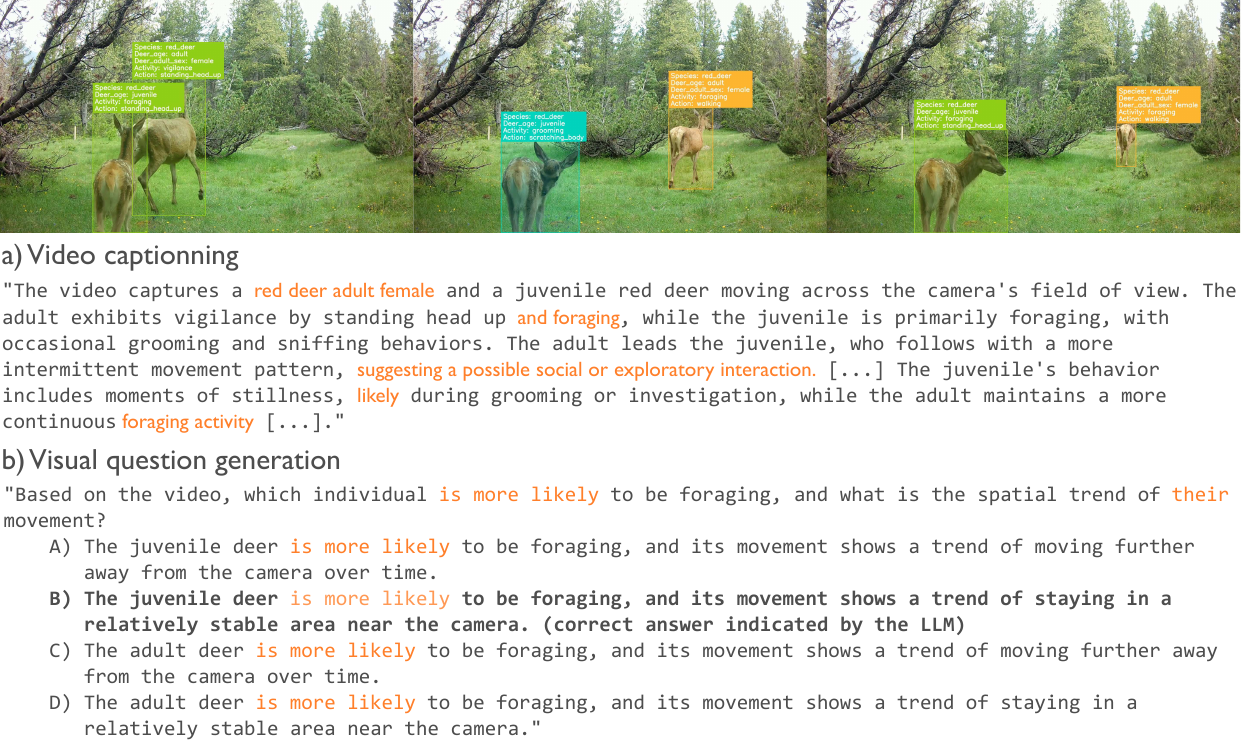}
    \caption{Exploratory use of Qwen3-8B to generate a) a video caption, and b) a visual question-answers pair given a ground-truth \texttt{.json} file only. Elements highlighted in orange are grammatically incorrect or prone to misinterpretation.
    In a), the adult maintains a more continuous movement than the juvenile, but not a more continuous activity. Since both individuals are equally likely to be foraging, the question in b) could be reformulated to ask about more differentiating activities such as vigilance.}
    \label{app-fig:extensions}
\end{figure}

From MammAlps-S2, we propose a TVR benchmark: Prompting-MammAlps. However the dataset could be used as the source to other benchmarks that would evaluate different reasoning capacities of VLMs. We show how one could use MammAlps-S2 for visual question answering and video captioning in \cref{app-fig:extensions}. We prompted Qwen3-8B to generate either a caption or a question with possible answers, given a subsampled \texttt{.json} annotation file only as input. While the generated text has some ecological or grammatical mistakes that would require manual correction, the process shows how the dense annotations of MammAlps-S2 can guide the development of such benchmarks without processing the videos. A related video question-answering benchmark is discussed in Mamooler et. al~\cite{mamooler2025finetuning}.